\documentclass[10pt,twocolumn,letterpaper]{article}

\usepackage[pagenumbers]{cvpr} %

\usepackage{amsmath,amssymb,amsthm}
\usepackage{graphicx,booktabs,multirow,tabularx}
\usepackage{xcolor}
\usepackage{colortbl}
\usepackage{tcolorbox}
\usepackage{tikz}
\usetikzlibrary{arrows.meta,positioning,fit,backgrounds}
\usepackage{algorithm}
\usepackage{algpseudocode}
\newcommand{\method}{\textnormal{\textsc{PACE-WM}}}
\newcommand{\rgp}{\mathrm{RAMP}}
\newcommand{\R}{\mathbb{R}}

\definecolor{harprow}{RGB}{238,244,250}

\newenvironment{progressbox}{%
\begin{tcolorbox}[colback=white!98!black,colframe=white!30!black,
boxsep=1.1pt,top=6.75pt,left=5pt,right=5pt,bottom=5pt,after skip=0pt]
\setlength{\parindent}{0pt}\noindent\textbf{Max--min interpretation of RAMP}\par\smallskip
\hrule height 0.4pt\relax\smallskip\noindent\ignorespaces
}{\end{tcolorbox}}

\usepackage{cuted} %

\usepackage{currfile} %

\usepackage{caption} %

\definecolor{cvprblue}{rgb}{0.21,0.49,0.74}
\usepackage[pagebackref,breaklinks,colorlinks,allcolors=cvprblue]{hyperref}

\def\confName{CVPR}
\def\confYear{2027}

\title{PACE-WM: Parallel World Modeling with Action History\\for Visual Planning}
\author{%
  Haishan Liu\textsuperscript{1,2,8}\quad
  Shijie Lian\textsuperscript{3,2}\quad
  Bin Yu\textsuperscript{4,2}\quad
  Yao Zhang\textsuperscript{5,2}\quad
  Changti Wu\textsuperscript{6,2}\quad
  Fanping Liu\textsuperscript{1}\\
  Jianzheng Liu\textsuperscript{1}\quad
  Lei Xie\textsuperscript{1,$\dagger$}\quad
  Cong Huang\textsuperscript{2,7,$\dagger$}\quad
  Kai Chen\textsuperscript{2,7,8,$\dagger$}\\
  {\small\textsuperscript{1}Nanjing University\quad 
  \textsuperscript{2}Zhongguancun Academy}\quad
  {\small\textsuperscript{3}Huazhong University of Science and Technology}\\
  {\small\textsuperscript{4}Harbin Institute of Technology}\quad
  {\small\textsuperscript{5}Wuhan University}\quad
  {\small\textsuperscript{6}East China Normal University}\\
  {\small\textsuperscript{7}Zhongguancun Institute of Artificial Intelligence}\quad
  {\small\textsuperscript{8}DeepCybo}\quad
}

\begin{document}
\maketitle
\begingroup
\renewcommand{\thefootnote}{$\dagger$}
\footnotetext{Corresponding author.}
\renewcommand{\thefootnote}{}
\footnotetext{Work done during internship at DeepCybo.}
\endgroup
\begin{strip}
\vspace{-47pt}
\centering
\includegraphics[width=\textwidth,height=.27844\textwidth]{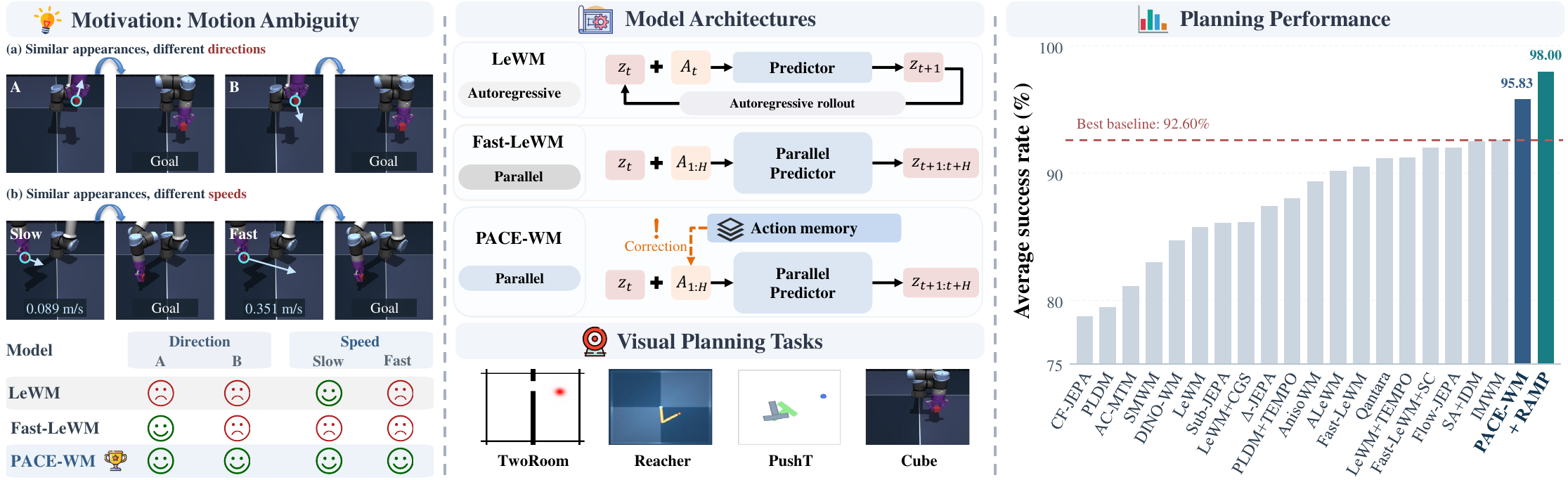}
\captionsetup{hypcap=false}
\captionof{figure}{\textbf{Action history for parallel visual planning.} Left: similar appearances conceal different motion directions or speeds. PACE-WM succeeds in both states of each pair, while each baseline fails in at least one. Middle: PACE-WM combines action memory and parallel prediction across four visual planning tasks. Right: average success reaches $95.83\%$ for PACE-WM and $98.00\%$ with RAMP.}
\label{fig:average_performance}
\label{fig:motivation_performance_overview}
\end{strip}

\begingroup
\renewcommand{\endabstract}{\par\vspace{-6pt}}
\begin{abstract}
JEPA-style world models support visual planning by predicting future representations conditioned on future actions. However, similar observations can conceal different motion states, making temporal context essential for interpreting future action effects. We propose \textbf{PACE-WM}, a parallel world model that uses executed-action history to contextualize future prediction and planning. Specifically, PACE-WM maintains a recurrent memory of executed actions and uses this memory to condition a causal Transformer. PACE-WM predicts multiple future representations in parallel without recursive predicted-state feedback within each prediction window. We further introduce Reference-Anchored Minimum Progress (RAMP) for action selection during planning. RAMP scores candidate actions by their predicted goal improvement along direct and split paths relative to a shared reference. Across four visual planning tasks, PACE-WM with RAMP achieves an average success rate of $98\%$ and plans approximately $7$--$9\times$ faster than autoregressive LeWM under matched CEM settings. We also construct four controlled benchmarks to evaluate PACE-WM's capabilities in motion disambiguation, goal-relative motion, and goal adaptation. Project page is available at \url{https://github.com/LHS180/PACE-WM}.
\end{abstract}

\endgroup
\section{Introduction}
World models support visual planning by predicting how an environment responds to candidate actions. Joint-embedding predictive architectures (JEPA) predict visual representations without reconstructing pixels~\cite{assran2023ijepa,bardes2024vjepa}, providing a foundation for modeling future outcomes in latent space. Recent approaches extend representation prediction to action-conditioned dynamics, enabling goal-directed planning from visual observations~\cite{zhou2025dinowm,sobal2025pldm,maes2026lewm}. A planner compares predicted outcomes with a goal representation and replans after executing selected actions. Effective planning therefore requires representations that preserve action-relevant dynamics. Repeated evaluation of candidate sequences also makes prediction efficiency essential for closed-loop control.

Visual planning requires understanding how future actions interact with ongoing motion. Visually similar robot configurations can conceal different motion directions or speeds, making the underlying dynamic state ambiguous. The paired examples in Fig.~\ref{fig:average_performance} illustrate this ambiguity through opposing motion directions and different speeds. Executed-action history provides clues to these differences by recording how past controls have reinforced or opposed motion. This context can help distinguish the effects of future actions that visual appearance alone leaves ambiguous. While visual history also supports state estimation~\cite{hafner2019planet,zhou2025dinowm}, past actions offer complementary dynamic context that can be updated incrementally without encoding additional images. Beyond modeling dynamic context, visual planning also requires efficient evaluation and selection of candidate action sequences. First, autoregressive rollouts~\cite{maes2026lewm,zhao2026subjepa,zhang2026dwm} evaluate candidates through sequential model calls, increasing planning overhead. This motivates predicting multiple future representations in parallel~\cite{gao2026fastlewm,feng2026ppwm}. Second, direct prediction and prediction through an intermediate state may assign different goal progress to the same action sequence, motivating a planning criterion that accounts for both estimates. We therefore combine executed-action memory, parallel prediction, and goal-directed planning.

We propose \method{}, a parallel world model that uses a recurrent encoder to summarize executed actions into persistent memory. Conditioned on this memory and future action prefixes, \method{} predicts multiple future representations in parallel through a causal Transformer~\cite{vaswani2017attention}. During closed-loop planning, \method{} updates the action memory with executed actions. This context helps the model account for how ongoing motion influences the effects of subsequent actions. For action selection, we introduce \emph{Reference-Anchored Minimum Progress} (RAMP), which compares direct and split predictions in terms of goal improvement relative to a shared reference action sequence. At unit weight, minimizing its cost is equivalent to maximizing the smaller of the two predicted improvements. This provides CEM~\cite{deboer2005cem} with a goal-directed and conservative scoring rule that accounts for disagreement between prediction paths when ranking candidate action sequences.

\method{} achieves an average success rate of $95.83\%$ across four visual planning tasks, increasing to $98.00\%$ with RAMP, as shown in Fig.~\ref{fig:average_performance}. Under matched CEM settings, \method{} with RAMP plans approximately $7$--$9\times$ faster than autoregressive LeWM~\cite{maes2026lewm}. To examine the role of executed-action context beyond aggregate performance, we further construct four controlled benchmarks to evaluate motion disambiguation, goal-relative motion, and goal adaptation. We demonstrate the benefits of action memory through comparisons with existing world models and independently trained visual-history and no-history variants.

Our contributions are threefold:
\begin{itemize}
\item We propose \method{}, which combines recurrent executed-action memory with causal parallel prediction to model future action effects in their dynamic context and support efficient visual planning.
\item We introduce RAMP, a reference-anchored planning criterion that maximizes the smaller goal improvement predicted by direct and split paths at unit weight.
\item We construct four controlled benchmarks to evaluate motion-dependent planning and validate our approach through experiments on four visual planning tasks, component ablations, and planning efficiency analyses.
\end{itemize}

\section{Related Work}
\par\addvspace{0.5\baselineskip}
\noindent\textbf{JEPA-style world models for visual planning.} Joint-embedding prediction learns visual representations without pixel reconstruction~\cite{assran2023ijepa,bardes2024vjepa}. Action-conditioned latent dynamics extend this approach to visual goal-directed planning~\cite{assran2025vjepa2,zhou2025dinowm,sobal2025pldm,maes2026lewm}. Representation learning spans pretrained visual features~\cite{oquab2023dinov2,zhou2025dinowm} and jointly learned encoders~\cite{maes2026lewm}, with anti-collapse regularization providing a foundation for learning nontrivial embeddings~\cite{bardes2022vicreg,zbontar2021barlow}. Subsequent work improves the structure of latent representations~\cite{zhao2026subjepa,achituve2026alewm,kang2026anisowm,byrd2026cfjepa,zhang2026dwm} and their sensitivity to actions~\cite{ivashkov2026smwm,zhang2026deltajepa,boylan2026acmtm,liu2026physicaljepa,fu2026cgs}. Alongside these representation advances, recent methods~\cite{gao2026fastlewm,feng2026ppwm} predict multiple future representations in parallel, reducing the sequential computation required by autoregressive rollouts. PACE-WM brings executed-action context into parallel latent prediction, allowing future action sequences to be evaluated in light of past controls.

\par\addvspace{0.5\baselineskip}
\noindent\textbf{History-conditioned dynamics.} Recurrent and autoregressive world models integrate past observations and actions for control~\cite{hafner2019planet,hafner2020dreamer,micheli2023iris}. Recent approaches use image-action histories to infer motion and slowly varying disturbances~\cite{jiang2026flowmo}, develop efficient visual-history mechanisms~\cite{guo2026history}, or condition parallel prediction on temporal differences between visual representations~\cite{feng2026ppwm}. These methods highlight the role of temporal context in predicting future dynamics. PACE-WM maintains a recurrent memory of executed actions to contextualize parallel prediction, providing a complementary source of dynamic information without encoding historical images.

\par\addvspace{0.5\baselineskip}
\noindent\textbf{Planning objectives and consistency.} Model-based control methods~\cite{chua2018pets,hansen2024tdmpc2} use learned dynamics to guide action selection, with planning efficiency improved through search optimization~\cite{pinneri2020icem,psenka2026grasp} and amortized action mappings~\cite{sun2026intact,nguyen2026gcidm}. Beyond latent-space proximity to the goal, temporal geometry~\cite{moukheiber2026tempo,yuan2026dtrc} guides planning and policy learning, while trajectory costs and demonstration-informed scores~\cite{huang2026trajlewm,gao2026imwm} refine candidate evaluation. Consistency-based methods~\cite{gao2026fastlewm,seo2026acid,wang2026aca} constrain prediction paths or action consequences, while D-JEPA~\cite{liu2026djepa} learns decision alignment from executed outcomes. RAMP compares goal progress supported by direct and split predictions relative to a shared reference, linking prediction-path differences to goal-directed action selection.

\section{Method}
\label{sec:method}
\begin{figure}[t]
\centering
\includegraphics[width=\columnwidth]{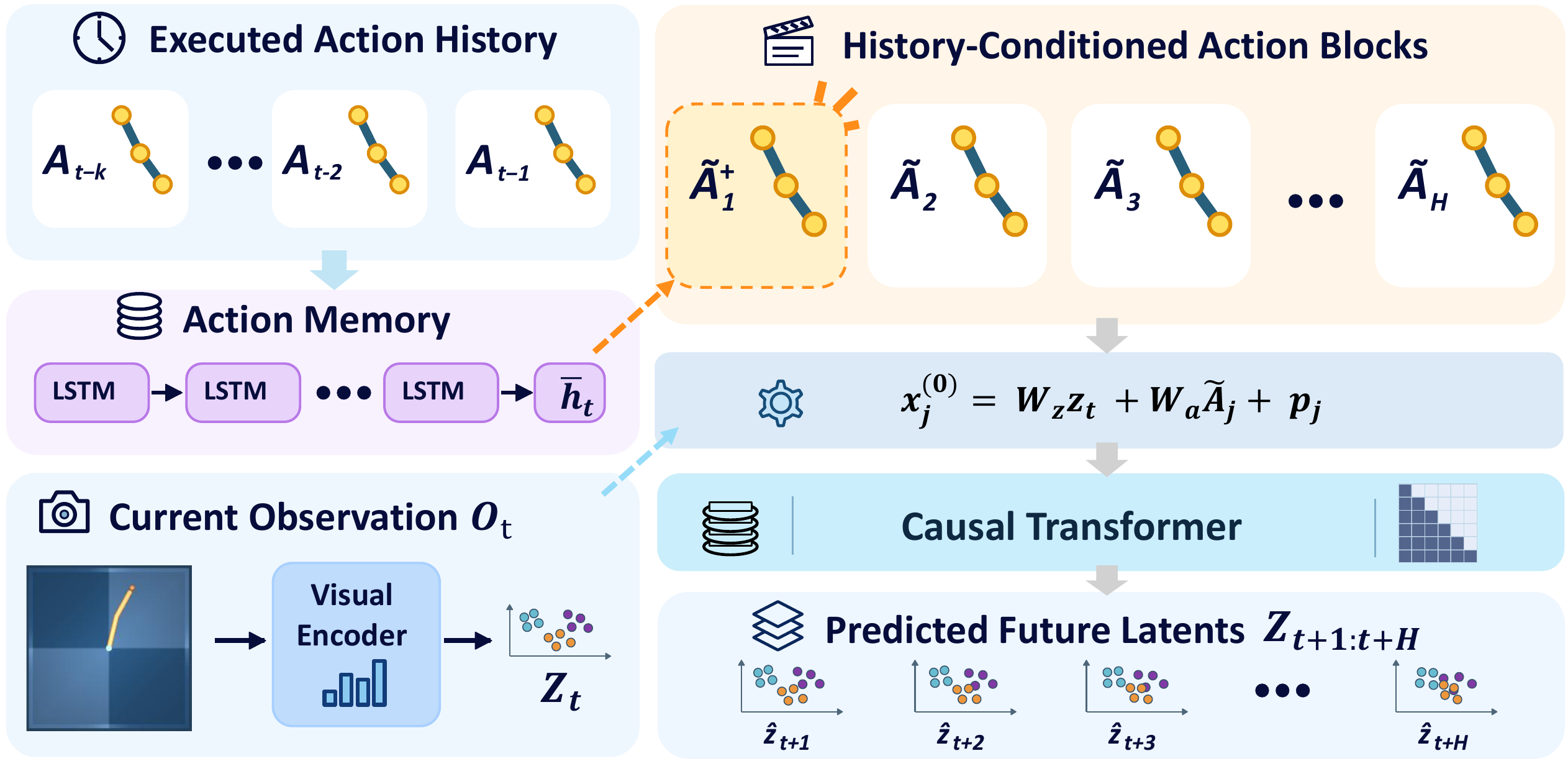}
\caption{\textbf{PACE-WM architecture.} Recurrent action memory adjusts the first future action block. A causal Transformer propagates this context across action tokens and uses the current visual representation to predict future latents in parallel.}
\label{fig:architecture}
\end{figure}

\method{} combines executed-action memory, parallel latent prediction, and goal-directed planning, as illustrated in Fig.~\ref{fig:architecture}. We summarize past controls in a recurrent memory and use this context to condition a causal Transformer that predicts multiple future representations in parallel. At planning time, RAMP evaluates candidate action sequences by their goal progress under direct and split predictions relative to a shared reference.

\subsection{Problem formulation}
In offline visual goal-conditioned planning, training data consist of observation--action trajectories $\mathcal D=\{\tau^{(n)}\}_{n=1}^{N}$ without reward annotations. Each trajectory is $\tau=(o_0,a_0,\ldots,a_{T-1},o_T)$, where $o_t$ is an image and $a_t\in\mathbb R^{d_a}$ is a primitive control. A shared encoder maps the current image $o_t$ and goal image $o_g$ to $z_t=f_\phi(o_t)\in\R^d$ and $z_g=f_\phi(o_g)$. A candidate action sequence $A=A_{1:H}$ contains $H$ blocks of $K$ consecutive controls:
\begin{equation}
 A_j=\operatorname{vec}(a_{t+(j-1)K:t+jK-1}),\quad 1\leq j\leq H.
\end{equation}
Given the current representation, available history, and candidate actions $A$, the world model predicts a future representation $\widehat z_j$ after each action block, approximating $f_\phi(o_{t+jK})$. For a fixed goal and context, planning selects the action sequence that minimizes a candidate cost $J(A)$:
\begin{equation}
 A^\star=\arg\min_A J(A).
 \label{eq:planning_problem}
\end{equation}
In closed-loop planning, the controller executes a sequence of planned actions and uses the resulting observation to update the next plan.

\subsection{Executed-action memory}
\label{sec:memory}
Executed actions provide temporal context for interpreting future control effects. For example, similar arm configurations can conceal opposing velocities, causing the same future controls to produce different outcomes. We encode past controls in a recurrent memory to provide the predictor with this dynamic context.

To retain temporal order with incremental updates, we instantiate the action-history encoder with an LSTM~\cite{hochreiter1997lstm} that summarizes executed action blocks $U_i$:
\begin{equation}
 (h_{i+1},c_{i+1})=\operatorname{LSTM}_{\psi}
 ([\operatorname{vec}(U_i),v_i],(h_i,c_i)),
 \label{eq:memory}
\end{equation}
where $h_i$ and $c_i$ are the hidden and cell states, and the validity mask $v_i$ distinguishes padding from executed zero-valued controls. Let $\bar h_t$ denote the hidden state used for prediction, which incorporates all executed controls, including any incomplete action block. We linearly project it into an additive correction for the predictor input:
\begin{equation}
 \delta_t=W_h\bar h_t+b_h.
 \label{eq:history_readout}
\end{equation}
We apply this correction only to the first future action block supplied to the predictor:
\begin{equation}
 \begin{aligned}
 \widetilde A_1&=A_1+\delta_t,\\
 \widetilde A_j&=A_j,\qquad j=2,\ldots,H.
 \end{aligned}
 \label{eq:injection}
\end{equation}
Applied before action projection, the history correction propagates through causal attention to inform all future predictions without adding memory tokens or altering executed controls. We initialize the readout to zero and enforce zero correction when no history is available. Let $m_t$ denote the action memory maintained from executed controls. During training, we compute this memory from controls executed before the sampled observation within the same episode. During planning, all candidates share the current executed-action memory $m_t$. Only controls executed in the environment update the persistent memory, which is reset at episode boundaries.

\subsection{Parallel latent prediction}
\method{} predicts multiple future latent representations in parallel, conditioned on the current visual representation, candidate actions, and executed-action memory:
\begin{equation}
 \widehat Z_{1:H}=F_\theta(z_t,A_{1:H};m_t).
 \label{eq:interface}
\end{equation}
The prediction after block $j$ depends on the action prefix $A_{1:j}$. We enforce this causal structure through masked attention while computing all future predictions in parallel. We combine the current latent, a history-conditioned action block, and a positional encoding into each token:
\begin{equation}
 x_j^{(0)}=W_z z_t+W_a\widetilde A_j+p_j,
 \label{eq:tokens}
\end{equation}
where $W_z$ and $W_a$ are learned projections and $p_j$ is a positional encoding. The term $W_z z_t$ provides the same visual context to each token. The Transformer processes the token sequence $X^{(0)}=(x_1^{(0)},\ldots,x_H^{(0)})$, and a shared head maps each output token to a future latent representation:
\begin{equation}
 \begin{aligned}
 X^{(\ell+1)}&=\mathcal T_\ell(X^{(\ell)};M),\quad 0\leq\ell<L,\\
 M_{ij}&=\begin{cases}
 0, & j\leq i,\\
 -\infty, & j>i.
 \end{cases}
 \end{aligned}
\end{equation}
\begin{equation}
 \widehat z_j=g_\theta(\operatorname{LN}(x_j^{(L)})).
 \label{eq:output}
\end{equation}
Causal attention propagates the injected history across future action prefixes. Interactions among hidden tokens capture temporal dependencies, allowing all $H$ future representations to be predicted in one forward pass without recursive predicted-state feedback.

\par\addvspace{0.5\baselineskip}
\noindent\textbf{Learning objective.} We jointly train the visual encoder, action memory, and predictor to match predicted latents with the representations of future observations:
\begin{align}
 \mathcal L_{\rm pred}
 &=\frac{1}{BHd}\sum_{b=1}^{B}\sum_{j=1}^{H}
 \|\widehat z_{b,j}-f_\phi(o_{b,t+jK})\|^2,\\
 \mathcal L&=\mathcal L_{\rm pred}+\beta\mathcal L_{\rm SIGReg},
 \label{eq:train}
\end{align}
where $B$ is the batch size and $\beta$ weights the SIGReg~\cite{balestriero2025lejepa} regularizer, which discourages representation collapse.

\subsection{Reference-Anchored Minimum Progress}
\label{sec:rgp}
\begin{figure}[t]
\centering
\includegraphics[width=\columnwidth]{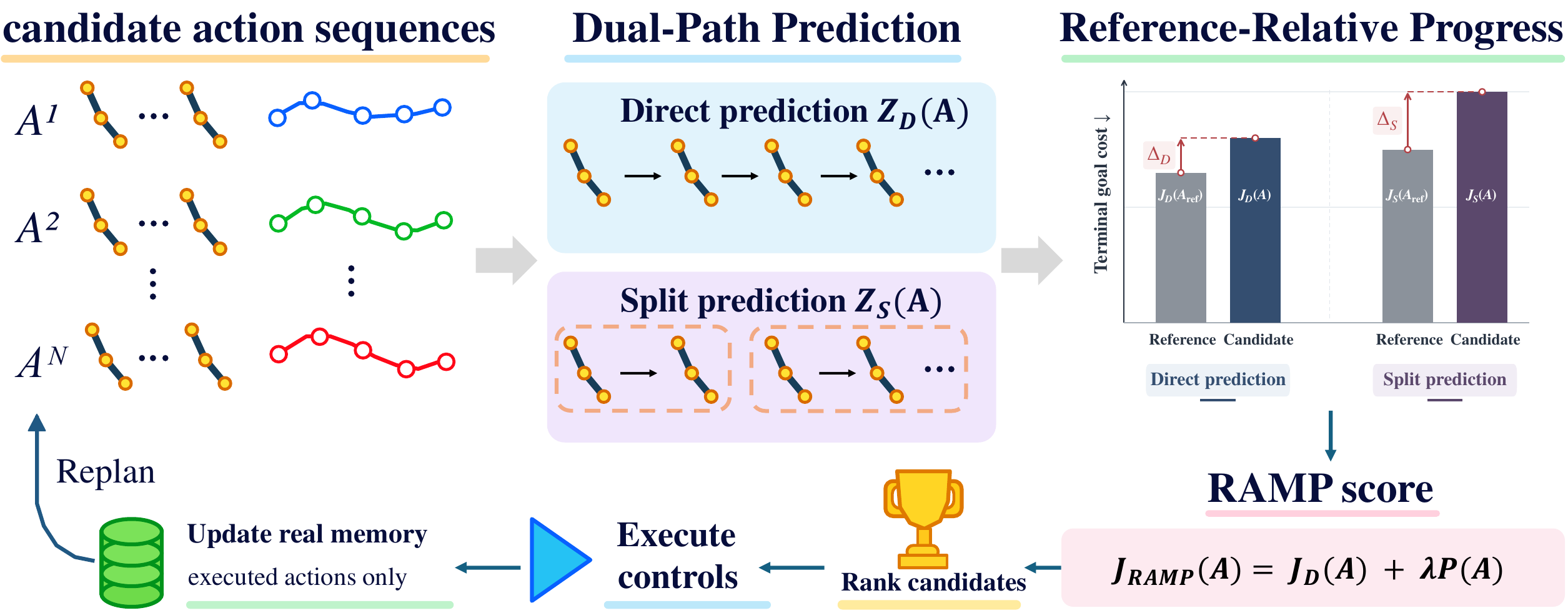}
\caption{\textbf{Closed-loop planning with RAMP.} RAMP evaluates candidate actions using the goal progress that both direct and split predictions support relative to a shared reference. CEM uses these scores to refine the plan, and executed actions update memory for subsequent planning. Bars schematically show path-specific terminal costs for the same reference and candidate sequences.}
\label{fig:ramp_planning}
\end{figure}

\begin{table*}[t]
\centering\small
\setlength{\tabcolsep}{4pt}
\renewcommand{\arraystretch}{0.94}
\setlength{\aboverulesep}{1pt}
\setlength{\belowrulesep}{1pt}
\caption{\textbf{Visual planning success rates (\%).} Bold/underline mark the best/second-best results. Avg. is the four-task mean. DINO-WM and PLDM scores are from LeWM~\cite{maes2026lewm}, and other baselines from their original papers under the source protocols. ALeWM reports mean $\pm$ SEM (standard error of the mean). Our results report mean $\pm$ sample SD over three training seeds.}
\label{tab:versions}
\begin{tabularx}{\textwidth}{@{}l*{5}{>{\raggedleft\arraybackslash}X}@{}}\toprule
\textbf{Model} & \textbf{TwoRoom} & \textbf{Reacher} & \textbf{PushT} & \textbf{Cube} & \textbf{Avg.}\\\midrule
PLDM~\cite{sobal2025pldm} & $97.00$ & $78.00$ & $78.00$ & $65.00$ & $79.50$\\
SMWM~\cite{ivashkov2026smwm} & $99.00$ & $66.00$ & $83.00$ & $84.00$ & $83.00$\\
DINO-WM~\cite{zhou2025dinowm} & $\mathbf{100.00}$ & $79.00$ & $74.00$ & $86.00$ & $84.75$\\
LeWM~\cite{maes2026lewm} & $87.00$ & $86.00$ & $96.00$ & $74.00$ & $85.75$\\
AnisoWM~\cite{kang2026anisowm} & $92.70$ & $88.70$ & $\underline{96.70}$ & $79.30$ & $89.35$\\
Fast-LeWM~\cite{gao2026fastlewm} & $98.00$ & $88.00$ & $96.00$ & $80.00$ & $90.50$\\
Fast-LeWM + SC~\cite{gao2026fastlewm} & $98.00$ & $90.00$ & $\mathbf{98.00}$ & $82.00$ & $92.00$\\
Flow-JEPA~\cite{huo2026flowjepa} & $\mathbf{100.00}$ & $90.00$ & $96.00$ & $82.00$ & $92.00$\\
SA+IDM~\cite{liu2026physicaljepa} & $\mathbf{100.00}$ & $85.00$ & $\mathbf{98.00}$ & $87.00$ & $92.50$\\
\midrule
CF-JEPA~\cite{byrd2026cfjepa} & $95.00{\scriptstyle\,\pm 4.00}$ & $61.00{\scriptstyle\,\pm 3.00}$ & $78.00{\scriptstyle\,\pm 4.00}$ & $81.00{\scriptstyle\,\pm 3.00}$ & $78.75$\\
AC-MTM~\cite{boylan2026acmtm} & $90.70{\scriptstyle\,\pm 0.60}$ & $68.30{\scriptstyle\,\pm 3.10}$ & $86.70{\scriptstyle\,\pm 1.50}$ & $78.80{\scriptstyle\,\pm 1.70}$ & $81.13$\\
Sub-JEPA~\cite{zhao2026subjepa} & $95.00{\scriptstyle\,\pm 2.76}$ & $84.00{\scriptstyle\,\pm 4.00}$ & $89.00{\scriptstyle\,\pm 5.33}$ & $76.33{\scriptstyle\,\pm 5.99}$ & $86.08$\\
LeWM + CGS~\cite{fu2026cgs} & $95.30{\scriptstyle\,\pm 1.20}$ & $80.70{\scriptstyle\,\pm 3.10}$ & $92.00{\scriptstyle\,\pm 3.50}$ & $76.70{\scriptstyle\,\pm 9.50}$ & $86.18$\\
$\Delta$-JEPA~\cite{zhang2026deltajepa} & $\mathbf{100.00}{\scriptstyle\,\pm 0.00}$ & $81.33{\scriptstyle\,\pm 0.50}$ & $89.07{\scriptstyle\,\pm 1.90}$ & $79.27{\scriptstyle\,\pm 1.81}$ & $87.42$\\
PLDM + TEMPO~\cite{moukheiber2026tempo} & $98.90{\scriptstyle\,\pm 0.30}$ & $94.50{\scriptstyle\,\pm 1.00}$ & $90.10{\scriptstyle\,\pm 1.20}$ & $68.70{\scriptstyle\,\pm 1.40}$ & $88.05$\\
ALeWM~\cite{achituve2026alewm} & $\mathbf{100.00}{\scriptstyle\,\pm 0.00}$ & $85.83{\scriptstyle\,\pm 1.01}$ & $96.00{\scriptstyle\,\pm 0.00}$ & $79.00{\scriptstyle\,\pm 0.76}$ & $90.21$\\
Qantara~\cite{rakhimov2026qantara} & $\mathbf{100.00}{\scriptstyle\,\pm 0.00}$ & $80.90{\scriptstyle\,\pm 1.80}$ & $90.10{\scriptstyle\,\pm 1.10}$ & $93.70{\scriptstyle\,\pm 0.70}$ & $91.18$\\
LeWM + TEMPO~\cite{moukheiber2026tempo} & $97.20{\scriptstyle\,\pm 0.60}$ & $\underline{95.50}{\scriptstyle\,\pm 0.70}$ & $95.10{\scriptstyle\,\pm 0.60}$ & $77.10{\scriptstyle\,\pm 1.70}$ & $91.23$\\
IMWM~\cite{gao2026imwm} & $\underline{99.20}{\scriptstyle\,\pm 1.00}$ & $83.80{\scriptstyle\,\pm 5.70}$ & $92.70{\scriptstyle\,\pm 3.40}$ & $\underline{94.70}{\scriptstyle\,\pm 1.90}$ & $92.60$\\
\midrule
\rowcolor{harprow}
\method{} & $\mathbf{100.00}{\scriptstyle\,\pm 0.00}$ & $92.67{\scriptstyle\,\pm 1.15}$ & $96.67{\scriptstyle\,\pm 2.31}$ & $94.00{\scriptstyle\,\pm 2.00}$ & $\underline{95.83}$\\
\rowcolor{harprow}
\method{} + RAMP & $\mathbf{100.00}{\scriptstyle\,\pm 0.00}$ & $\mathbf{98.00}{\scriptstyle\,\pm 0.00}$ & $\mathbf{98.00}{\scriptstyle\,\pm 2.00}$ & $\mathbf{96.00}{\scriptstyle\,\pm 0.00}$ & $\mathbf{98.00}$\\
\bottomrule\end{tabularx}

\end{table*}

Direct and split predictions can assign different goal progress to the same candidate action sequence. We introduce RAMP as a conservative planning criterion that accounts for this disagreement by comparing each path's predicted progress relative to a shared reference. At unit weight, it favors the smaller of the two predicted improvements, as illustrated in Fig.~\ref{fig:ramp_planning}.

\par\addvspace{0.5\baselineskip}
\noindent\textbf{Direct and split predictions.} For the same candidate $A$, the direct prediction is $z_D(A)=F_\theta(z_t,A;m_t)_H$. For a split point $1\leq k<H$, the split path reuses the direct prediction $\widehat z_k$ after $k$ action blocks. It advances a private memory copy with the same candidate prefix, then predicts the remaining blocks from this intermediate state:
\begin{align}
 m_t^A&=\operatorname{LSTM}_{\psi}(A_{1:k};m_t),\\
 z_S(A)&=F_\theta(\widehat z_k,A_{k+1:H};m_t^A)_{H-k},
 \label{eq:split}
\end{align}
where $\operatorname{LSTM}_{\psi}$ applies the memory update from Sec.~\ref{sec:memory} sequentially to the copied state. The suffix restarts positional encoding and first-block history conditioning. Both paths share model weights and the intermediate prediction, so their errors need not be independent. Their terminal goal costs are:
\begin{equation}
 \begin{aligned}
 J_D(A)&=\|z_D(A)-z_g\|_2^2,\\
 J_S(A)&=\|z_S(A)-z_g\|_2^2.
 \end{aligned}
 \label{eq:path_goal_costs}
\end{equation}

\par\addvspace{0.5\baselineskip}
\noindent\textbf{Reference-relative progress.} At CEM iteration $i$, we use the sampling-distribution mean $A_{\rm ref}=\mu^{(i)}$ as the reference, anchoring comparisons to the current search center. We include this sequence in the candidate batch and evaluate it through both paths using the same $z_t$ and $m_t$ as the other candidates. The paths share the reference actions but compute separate reference costs. Each measures a candidate's improvement relative to its own reference cost:
\begin{equation}
 \Delta_p(A)=J_p(A_{\rm ref})-J_p(A),\quad p\in\{D,S\}.
\end{equation}
Positive $\Delta_p(A)$ indicates predicted improvement over the reference. Reference subtraction cancels additive, candidate-independent cost offsets within each path. RAMP penalizes the amount by which direct-path progress exceeds split-path progress:
\begin{equation}
 P(A)=[\Delta_D(A)-\Delta_S(A)]_+,\quad [x]_+=\max(0,x).
 \label{eq:penalty}
\end{equation}
The planning cost is defined as follows:
\begin{equation}
 J_{\rgp}(A)=J_D(A)+\lambda P(A),
 \label{eq:rgpcost}
\end{equation}
where $\lambda\geq0$ controls the penalty strength. Setting $\lambda=0$ recovers direct goal-distance scoring.

\begin{progressbox}
For a fixed reference and $\lambda=1$,
{\small
\setlength{\medmuskip}{2mu}
\setlength{\thickmuskip}{3mu}
\begin{equation}
 J_{\rgp}(A)=J_D(A_{\rm ref})-\min\{\Delta_D(A),\Delta_S(A)\}.
 \label{eq:maxmin}
\end{equation}
}%
Thus, minimizing the RAMP cost is equivalent to maximizing the smaller predicted improvement. This equivalence characterizes the scoring rule, not a bound on progress in the environment.
\par\smallskip
The proof is provided in Appendix~\ref{supp-app:theory}.
\end{progressbox}

\par\addvspace{0.5\baselineskip}
\noindent\textbf{Closed-loop execution.} CEM scores candidate sequences using Eq.~\eqref{eq:rgpcost} and updates its sampling distribution from the elites. The controller executes the mean action sequence of the final distribution and updates its memory with the executed controls. If the task remains unfinished and the control budget allows, it replans from the new observation.

\section{Experiments}
\label{sec:experiments}

We evaluate PACE-WM's planning performance and efficiency across four visual control tasks. To examine the role of executed-action memory, we use four controlled benchmarks covering motion disambiguation, goal-relative motion, and goal adaptation. We then analyze how RAMP affects action selection, followed by component ablations and parameter studies that assess module contributions and sensitivity to key settings. Additional experimental results and analyses are provided in Appendix~\ref{supp-app:experiments}.

\subsection{Experimental Setup}
\label{sec:experimental_setup}
\par\addvspace{0.5\baselineskip}
\noindent\textbf{Tasks and metrics.} We evaluate visual goal reaching on TwoRoom~\cite{sobal2025pldm}, Reacher~\cite{tassa2018control}, PushT~\cite{chi2023diffusionpolicy}, and OG-Bench Cube~\cite{park2024ogbench}.\ We also construct four controlled Reacher benchmarks to examine the role of executed-action history in motion disambiguation, goal-relative motion, and goal adaptation, as detailed in Sec.~\ref{sec:history_benchmarks}. Performance is measured by success rate under each task's success criterion.

\begin{table*}[t]
\centering
\small
\setlength{\tabcolsep}{3pt}
\renewcommand{\arraystretch}{1.02}
\setlength{\aboverulesep}{1.5pt}
\setlength{\belowrulesep}{1.5pt}
\setlength{\arrayrulewidth}{0.3pt}
\caption{\textbf{Visual planning on four controlled Reacher benchmarks.} Success rates (\%, mean $\pm$ sample SD over three training seeds) using 25-step histories and Base planning, evaluated at 25/50 controls with replanning after 25. Avg. weights benchmarks equally after averaging their conditions. Bold/underline mark the best/second-best results per row.}
\label{tab:history_benchmarks}
\begin{tabularx}{\textwidth}{@{}l|lc|*{4}{>{\raggedleft\arraybackslash}X}>{\columncolor{harprow}[\tabcolsep][0pt]\raggedleft\arraybackslash}X@{}}
\toprule
\textbf{Benchmark} & \textbf{Condition} & \textbf{Controls} & \multicolumn{1}{c}{LeWM} & \multicolumn{1}{c}{Fast-LeWM} & \multicolumn{1}{c}{w/o history} & \multicolumn{1}{c}{Visual history} & \multicolumn{1}{c}{\cellcolor{harprow}\method{}} \\
\midrule
\multirow{4}{*}{\textbf{(a) Direction ambiguity}} & \multirow{2}{*}{Replayed history} & 25 & \makebox[2.9em][r]{62.33}{\scriptsize\makebox[0.9em][r]{$\pm$}\makebox[2.6em][l]{0.58}} & \makebox[2.9em][r]{64.33}{\scriptsize\makebox[0.9em][r]{$\pm$}\makebox[2.6em][l]{4.04}} & \makebox[2.9em][r]{59.33}{\scriptsize\makebox[0.9em][r]{$\pm$}\makebox[2.6em][l]{2.52}} & \makebox[2.9em][r]{\underline{75.33}}{\scriptsize\makebox[0.9em][r]{$\pm$}\makebox[2.6em][l]{4.04}} & \makebox[2.9em][r]{\textbf{96.67}}{\scriptsize\makebox[0.9em][r]{$\pm$}\makebox[2.6em][l]{2.08}} \\
 &  & 50 & \makebox[2.9em][r]{91.67}{\scriptsize\makebox[0.9em][r]{$\pm$}\makebox[2.6em][l]{1.53}} & \makebox[2.9em][r]{90.33}{\scriptsize\makebox[0.9em][r]{$\pm$}\makebox[2.6em][l]{2.08}} & \makebox[2.9em][r]{\underline{93.33}}{\scriptsize\makebox[0.9em][r]{$\pm$}\makebox[2.6em][l]{1.53}} & \makebox[2.9em][r]{91.67}{\scriptsize\makebox[0.9em][r]{$\pm$}\makebox[2.6em][l]{2.52}} & \makebox[2.9em][r]{\textbf{99.00}}{\scriptsize\makebox[0.9em][r]{$\pm$}\makebox[2.6em][l]{1.00}} \\
\cmidrule(l){2-8}
 & \multirow{2}{*}{Generated history} & 25 & \makebox[2.9em][r]{63.33}{\scriptsize\makebox[0.9em][r]{$\pm$}\makebox[2.6em][l]{4.04}} & \makebox[2.9em][r]{63.33}{\scriptsize\makebox[0.9em][r]{$\pm$}\makebox[2.6em][l]{3.51}} & \makebox[2.9em][r]{63.00}{\scriptsize\makebox[0.9em][r]{$\pm$}\makebox[2.6em][l]{3.61}} & \makebox[2.9em][r]{\underline{63.67}}{\scriptsize\makebox[0.9em][r]{$\pm$}\makebox[2.6em][l]{2.52}} & \makebox[2.9em][r]{\textbf{96.00}}{\scriptsize\makebox[0.9em][r]{$\pm$}\makebox[2.6em][l]{1.73}} \\
 &  & 50 & \makebox[2.9em][r]{88.67}{\scriptsize\makebox[0.9em][r]{$\pm$}\makebox[2.6em][l]{2.08}} & \makebox[2.9em][r]{\underline{92.33}}{\scriptsize\makebox[0.9em][r]{$\pm$}\makebox[2.6em][l]{2.08}} & \makebox[2.9em][r]{91.33}{\scriptsize\makebox[0.9em][r]{$\pm$}\makebox[2.6em][l]{2.31}} & \makebox[2.9em][r]{90.00}{\scriptsize\makebox[0.9em][r]{$\pm$}\makebox[2.6em][l]{2.00}} & \makebox[2.9em][r]{\textbf{99.67}}{\scriptsize\makebox[0.9em][r]{$\pm$}\makebox[2.6em][l]{0.58}} \\
\midrule
\multirow{4}{*}{\textbf{(b) Speed ambiguity}} & Slow & 25 & \makebox[2.9em][r]{5.33}{\scriptsize\makebox[0.9em][r]{$\pm$}\makebox[2.6em][l]{4.62}} & \makebox[2.9em][r]{5.33}{\scriptsize\makebox[0.9em][r]{$\pm$}\makebox[2.6em][l]{2.31}} & \makebox[2.9em][r]{8.00}{\scriptsize\makebox[0.9em][r]{$\pm$}\makebox[2.6em][l]{4.00}} & \makebox[2.9em][r]{\underline{92.00}}{\scriptsize\makebox[0.9em][r]{$\pm$}\makebox[2.6em][l]{6.93}} & \makebox[2.9em][r]{\textbf{97.33}}{\scriptsize\makebox[0.9em][r]{$\pm$}\makebox[2.6em][l]{4.62}} \\
 & Fast & 25 & \makebox[2.9em][r]{\underline{0.00}}{\scriptsize\makebox[0.9em][r]{$\pm$}\makebox[2.6em][l]{0.00}} & \makebox[2.9em][r]{\underline{0.00}}{\scriptsize\makebox[0.9em][r]{$\pm$}\makebox[2.6em][l]{0.00}} & \makebox[2.9em][r]{\underline{0.00}}{\scriptsize\makebox[0.9em][r]{$\pm$}\makebox[2.6em][l]{0.00}} & \makebox[2.9em][r]{\underline{0.00}}{\scriptsize\makebox[0.9em][r]{$\pm$}\makebox[2.6em][l]{0.00}} & \makebox[2.9em][r]{\textbf{2.67}}{\scriptsize\makebox[0.9em][r]{$\pm$}\makebox[2.6em][l]{4.62}} \\
\cmidrule(l){2-8}
 & Slow & 50 & \makebox[2.9em][r]{88.00}{\scriptsize\makebox[0.9em][r]{$\pm$}\makebox[2.6em][l]{4.00}} & \makebox[2.9em][r]{84.00}{\scriptsize\makebox[0.9em][r]{$\pm$}\makebox[2.6em][l]{0.00}} & \makebox[2.9em][r]{86.67}{\scriptsize\makebox[0.9em][r]{$\pm$}\makebox[2.6em][l]{2.31}} & \makebox[2.9em][r]{\underline{98.67}}{\scriptsize\makebox[0.9em][r]{$\pm$}\makebox[2.6em][l]{2.31}} & \makebox[2.9em][r]{\textbf{100.00}}{\scriptsize\makebox[0.9em][r]{$\pm$}\makebox[2.6em][l]{0.00}} \\
 & Fast & 50 & \makebox[2.9em][r]{53.33}{\scriptsize\makebox[0.9em][r]{$\pm$}\makebox[2.6em][l]{8.33}} & \makebox[2.9em][r]{57.33}{\scriptsize\makebox[0.9em][r]{$\pm$}\makebox[2.6em][l]{15.14}} & \makebox[2.9em][r]{46.67}{\scriptsize\makebox[0.9em][r]{$\pm$}\makebox[2.6em][l]{11.55}} & \makebox[2.9em][r]{\underline{58.67}}{\scriptsize\makebox[0.9em][r]{$\pm$}\makebox[2.6em][l]{8.33}} & \makebox[2.9em][r]{\textbf{96.00}}{\scriptsize\makebox[0.9em][r]{$\pm$}\makebox[2.6em][l]{4.00}} \\
\midrule
\multirow{4}{*}{\textbf{(c) Goal-relative motion}} & Toward & 25 & \makebox[2.9em][r]{84.00}{\scriptsize\makebox[0.9em][r]{$\pm$}\makebox[2.6em][l]{4.00}} & \makebox[2.9em][r]{78.67}{\scriptsize\makebox[0.9em][r]{$\pm$}\makebox[2.6em][l]{6.11}} & \makebox[2.9em][r]{78.67}{\scriptsize\makebox[0.9em][r]{$\pm$}\makebox[2.6em][l]{10.07}} & \makebox[2.9em][r]{\underline{85.33}}{\scriptsize\makebox[0.9em][r]{$\pm$}\makebox[2.6em][l]{4.62}} & \makebox[2.9em][r]{\textbf{96.00}}{\scriptsize\makebox[0.9em][r]{$\pm$}\makebox[2.6em][l]{4.00}} \\
 & Away & 25 & \makebox[2.9em][r]{48.00}{\scriptsize\makebox[0.9em][r]{$\pm$}\makebox[2.6em][l]{6.93}} & \makebox[2.9em][r]{38.67}{\scriptsize\makebox[0.9em][r]{$\pm$}\makebox[2.6em][l]{4.62}} & \makebox[2.9em][r]{42.67}{\scriptsize\makebox[0.9em][r]{$\pm$}\makebox[2.6em][l]{4.62}} & \makebox[2.9em][r]{\underline{62.67}}{\scriptsize\makebox[0.9em][r]{$\pm$}\makebox[2.6em][l]{2.31}} & \makebox[2.9em][r]{\textbf{96.00}}{\scriptsize\makebox[0.9em][r]{$\pm$}\makebox[2.6em][l]{4.00}} \\
\cmidrule(l){2-8}
 & Toward & 50 & \makebox[2.9em][r]{94.67}{\scriptsize\makebox[0.9em][r]{$\pm$}\makebox[2.6em][l]{2.31}} & \makebox[2.9em][r]{97.33}{\scriptsize\makebox[0.9em][r]{$\pm$}\makebox[2.6em][l]{2.31}} & \makebox[2.9em][r]{94.67}{\scriptsize\makebox[0.9em][r]{$\pm$}\makebox[2.6em][l]{6.11}} & \makebox[2.9em][r]{\underline{98.67}}{\scriptsize\makebox[0.9em][r]{$\pm$}\makebox[2.6em][l]{2.31}} & \makebox[2.9em][r]{\textbf{100.00}}{\scriptsize\makebox[0.9em][r]{$\pm$}\makebox[2.6em][l]{0.00}} \\
 & Away & 50 & \makebox[2.9em][r]{\underline{94.67}}{\scriptsize\makebox[0.9em][r]{$\pm$}\makebox[2.6em][l]{6.11}} & \makebox[2.9em][r]{81.33}{\scriptsize\makebox[0.9em][r]{$\pm$}\makebox[2.6em][l]{10.07}} & \makebox[2.9em][r]{90.67}{\scriptsize\makebox[0.9em][r]{$\pm$}\makebox[2.6em][l]{8.33}} & \makebox[2.9em][r]{\underline{94.67}}{\scriptsize\makebox[0.9em][r]{$\pm$}\makebox[2.6em][l]{2.31}} & \makebox[2.9em][r]{\textbf{100.00}}{\scriptsize\makebox[0.9em][r]{$\pm$}\makebox[2.6em][l]{0.00}} \\
\midrule
\multirow{6}{*}{\textbf{(d) Goal switching}} & $0^\circ$ (original) & 25 & \makebox[2.9em][r]{82.67}{\scriptsize\makebox[0.9em][r]{$\pm$}\makebox[2.6em][l]{5.03}} & \makebox[2.9em][r]{\underline{88.00}}{\scriptsize\makebox[0.9em][r]{$\pm$}\makebox[2.6em][l]{2.00}} & \makebox[2.9em][r]{85.33}{\scriptsize\makebox[0.9em][r]{$\pm$}\makebox[2.6em][l]{2.31}} & \makebox[2.9em][r]{87.33}{\scriptsize\makebox[0.9em][r]{$\pm$}\makebox[2.6em][l]{5.77}} & \makebox[2.9em][r]{\textbf{89.33}}{\scriptsize\makebox[0.9em][r]{$\pm$}\makebox[2.6em][l]{4.62}} \\
 & $90^\circ$ switch & 25 & \makebox[2.9em][r]{76.67}{\scriptsize\makebox[0.9em][r]{$\pm$}\makebox[2.6em][l]{5.03}} & \makebox[2.9em][r]{75.33}{\scriptsize\makebox[0.9em][r]{$\pm$}\makebox[2.6em][l]{4.62}} & \makebox[2.9em][r]{76.00}{\scriptsize\makebox[0.9em][r]{$\pm$}\makebox[2.6em][l]{2.00}} & \makebox[2.9em][r]{\underline{77.33}}{\scriptsize\makebox[0.9em][r]{$\pm$}\makebox[2.6em][l]{3.06}} & \makebox[2.9em][r]{\textbf{92.00}}{\scriptsize\makebox[0.9em][r]{$\pm$}\makebox[2.6em][l]{2.00}} \\
 & $180^\circ$ switch & 25 & \makebox[2.9em][r]{58.67}{\scriptsize\makebox[0.9em][r]{$\pm$}\makebox[2.6em][l]{4.16}} & \makebox[2.9em][r]{62.00}{\scriptsize\makebox[0.9em][r]{$\pm$}\makebox[2.6em][l]{5.29}} & \makebox[2.9em][r]{66.67}{\scriptsize\makebox[0.9em][r]{$\pm$}\makebox[2.6em][l]{2.31}} & \makebox[2.9em][r]{\underline{67.33}}{\scriptsize\makebox[0.9em][r]{$\pm$}\makebox[2.6em][l]{7.02}} & \makebox[2.9em][r]{\textbf{88.67}}{\scriptsize\makebox[0.9em][r]{$\pm$}\makebox[2.6em][l]{3.06}} \\
\cmidrule(l){2-8}
 & $0^\circ$ (original) & 50 & \makebox[2.9em][r]{94.67}{\scriptsize\makebox[0.9em][r]{$\pm$}\makebox[2.6em][l]{1.15}} & \makebox[2.9em][r]{\underline{97.33}}{\scriptsize\makebox[0.9em][r]{$\pm$}\makebox[2.6em][l]{1.15}} & \makebox[2.9em][r]{96.00}{\scriptsize\makebox[0.9em][r]{$\pm$}\makebox[2.6em][l]{2.00}} & \makebox[2.9em][r]{96.00}{\scriptsize\makebox[0.9em][r]{$\pm$}\makebox[2.6em][l]{2.00}} & \makebox[2.9em][r]{\textbf{99.33}}{\scriptsize\makebox[0.9em][r]{$\pm$}\makebox[2.6em][l]{1.15}} \\
 & $90^\circ$ switch & 50 & \makebox[2.9em][r]{94.00}{\scriptsize\makebox[0.9em][r]{$\pm$}\makebox[2.6em][l]{2.00}} & \makebox[2.9em][r]{88.00}{\scriptsize\makebox[0.9em][r]{$\pm$}\makebox[2.6em][l]{5.29}} & \makebox[2.9em][r]{89.33}{\scriptsize\makebox[0.9em][r]{$\pm$}\makebox[2.6em][l]{3.06}} & \makebox[2.9em][r]{\underline{95.33}}{\scriptsize\makebox[0.9em][r]{$\pm$}\makebox[2.6em][l]{2.31}} & \makebox[2.9em][r]{\textbf{99.33}}{\scriptsize\makebox[0.9em][r]{$\pm$}\makebox[2.6em][l]{1.15}} \\
 & $180^\circ$ switch & 50 & \makebox[2.9em][r]{86.67}{\scriptsize\makebox[0.9em][r]{$\pm$}\makebox[2.6em][l]{5.77}} & \makebox[2.9em][r]{90.67}{\scriptsize\makebox[0.9em][r]{$\pm$}\makebox[2.6em][l]{3.06}} & \makebox[2.9em][r]{90.67}{\scriptsize\makebox[0.9em][r]{$\pm$}\makebox[2.6em][l]{3.06}} & \makebox[2.9em][r]{\underline{94.00}}{\scriptsize\makebox[0.9em][r]{$\pm$}\makebox[2.6em][l]{4.00}} & \makebox[2.9em][r]{\textbf{98.67}}{\scriptsize\makebox[0.9em][r]{$\pm$}\makebox[2.6em][l]{1.15}} \\
\midrule
\multicolumn{2}{c}{} & 25 & \makebox[2.9em][r]{51.04}{\scriptsize\phantom{\makebox[0.9em][r]{$\pm$}\makebox[2.6em][l]{0.00}}} & \makebox[2.9em][r]{50.07}{\scriptsize\phantom{\makebox[0.9em][r]{$\pm$}\makebox[2.6em][l]{0.00}}} & \makebox[2.9em][r]{50.46}{\scriptsize\phantom{\makebox[0.9em][r]{$\pm$}\makebox[2.6em][l]{0.00}}} & \makebox[2.9em][r]{\underline{66.71}}{\scriptsize\phantom{\makebox[0.9em][r]{$\pm$}\makebox[2.6em][l]{0.00}}} & \makebox[2.9em][r]{\textbf{83.08}}{\scriptsize\phantom{\makebox[0.9em][r]{$\pm$}\makebox[2.6em][l]{0.00}}} \\
\multicolumn{2}{c}{\multirow{-2}{*}{\textbf{Average over four benchmarks}}} & 50 & \makebox[2.9em][r]{86.82}{\scriptsize\phantom{\makebox[0.9em][r]{$\pm$}\makebox[2.6em][l]{0.00}}} & \makebox[2.9em][r]{85.83}{\scriptsize\phantom{\makebox[0.9em][r]{$\pm$}\makebox[2.6em][l]{0.00}}} & \makebox[2.9em][r]{85.92}{\scriptsize\phantom{\makebox[0.9em][r]{$\pm$}\makebox[2.6em][l]{0.00}}} & \makebox[2.9em][r]{\underline{90.32}}{\scriptsize\phantom{\makebox[0.9em][r]{$\pm$}\makebox[2.6em][l]{0.00}}} & \makebox[2.9em][r]{\textbf{99.11}}{\scriptsize\phantom{\makebox[0.9em][r]{$\pm$}\makebox[2.6em][l]{0.00}}} \\
\bottomrule
\end{tabularx}

\end{table*}

\par\addvspace{0.5\baselineskip}
\noindent\textbf{Implementation details.} Our causal Transformer uses 6 layers, 6 attention heads, and a hidden dimension of 192. The action-memory LSTM has a hidden dimension of 32. We train with AdamW using a learning rate of $5\times10^{-5}$, weight decay of $10^{-3}$, and gradient clipping at 1. The SIGReg weight is $\beta=0.09$. Experiments run on NVIDIA RTX PRO 6000 Blackwell GPUs. Further implementation and planning details are provided in Appendix~\ref{supp-app:implementation}.

\par\addvspace{0.5\baselineskip}
\noindent\textbf{Planning protocol.} Unless otherwise specified, each CEM call optimizes 5 action blocks of 5 controls each over 30 iterations, with 300 candidates and 30 elites per iteration. The controller executes the resulting 25-control sequence and replans from a new image if needed, within a total budget of 50 controls. RAMP uses $\lambda=1$ and splits predictions after $k=2$ blocks. For planner comparisons, Base uses terminal latent goal distance alone, while SC~\cite{gao2026fastlewm} adds a latent self-consistency penalty.

\subsection{Visual Planning}
\label{sec:main_results}
To assess visual planning performance and the benefits of RAMP, we evaluate \method{} across four tasks. With Base planning, \method{} achieves an average success rate of $95.83\%$, exceeding the published baseline averages in Table~\ref{tab:versions}. RAMP further raises the average to $98.00\%$, yielding the best or joint-best results on all four tasks. It improves success by $5.33$ percentage points on Reacher, $2.00$ on Cube, and $1.33$ on PushT while maintaining $100.00\%$ on TwoRoom. These results demonstrate the strong visual planning performance of the base model and the additional gains from RAMP in selecting goal-directed actions.

\subsection{Planning with Executed-Action Memory}
\label{sec:history_benchmarks}
To assess the role of executed-action memory in motion-dependent planning, we construct four controlled Reacher benchmarks. We compare \method{} with LeWM, Fast-LeWM, and two independently trained variants: w/o history removes action memory, while Visual history uses an LSTM to encode past images instead of executed actions. All methods use Base planning to isolate the role of history. Evaluation cases are constructed using physical criteria and native-simulator validation. Results appear in Table~\ref{tab:history_benchmarks}, with construction details provided in Appendix~\ref{supp-app:benchmarks}.

\begin{table*}[t]
\centering\small
\setlength{\tabcolsep}{7pt}
\renewcommand{\arraystretch}{1.00}
\setlength{\aboverulesep}{1.5pt}
\setlength{\belowrulesep}{1.5pt}
\caption{\textbf{History and planning ablations.} Success rates (\%, mean $\pm$ sample SD over three training seeds) under the main-experiment protocol. (a) History representations with the parallel predictor architecture and RAMP planner fixed. (b) Planning criteria evaluated using the same trained models with LSTM-based action memory. Bold/underline mark the best/second-best results per task.}
\label{tab:combined_ablation}
\label{tab:history_ablation}
\label{tab:planning_ablation}
\begin{tabularx}{\textwidth}{@{}lll*{4}{>{\raggedleft\arraybackslash}X}@{}}
\toprule
History input & Encoder & Planner & TwoRoom & Reacher & PushT & Cube \\
\midrule
None & --- & Base & $95.33_{\pm1.15}$ & $82.67_{\pm3.06}$ & $94.67_{\pm3.06}$ & $92.00_{\pm2.00}$ \\
\midrule
\multicolumn{7}{l}{\textit{(a) History ablation}} \\
None & --- & RAMP & $97.33_{\pm1.15}$ & $84.00_{\pm3.46}$ & $96.67_{\pm1.15}$ & $94.00_{\pm2.00}$ \\
Actions & Linear & RAMP & $98.00_{\pm2.00}$ & $94.00_{\pm3.46}$ & $94.67_{\pm2.31}$ & $92.00_{\pm2.00}$ \\
Actions & GRU & RAMP & $\underline{98.67}_{\pm1.15}$ & $96.00_{\pm0.00}$ & $\underline{97.33}_{\pm2.31}$ & $\underline{95.33}_{\pm3.06}$ \\
Images & LSTM & RAMP & $97.33_{\pm1.15}$ & $86.00_{\pm5.29}$ & $93.33_{\pm3.06}$ & $93.33_{\pm1.15}$ \\
\midrule
\multicolumn{7}{l}{\textit{(b) Planning ablation}} \\
\rowcolor{harprow}
Actions & LSTM & Base & $\mathbf{100.00}_{\pm0.00}$ & $92.67_{\pm1.15}$ & $96.67_{\pm2.31}$ & $94.00_{\pm2.00}$ \\
Actions & LSTM & SC & $97.33_{\pm1.15}$ & $96.67_{\pm4.16}$ & $\underline{97.33}_{\pm3.06}$ & $94.67_{\pm1.15}$ \\
Actions & LSTM & Split-only & $98.00_{\pm2.00}$ & $95.33_{\pm1.15}$ & $91.33_{\pm2.31}$ & $91.33_{\pm1.15}$ \\
Actions & LSTM & Unanchored & $\underline{98.67}_{\pm1.15}$ & $\underline{97.33}_{\pm4.62}$ & $\mathbf{98.00}_{\pm2.00}$ & $92.00_{\pm2.00}$ \\
\midrule
\multicolumn{7}{l}{\textit{Complete model}} \\
\rowcolor{harprow}
Actions & LSTM & RAMP & $\mathbf{100.00}_{\pm0.00}$ & $\mathbf{98.00}_{\pm0.00}$ & $\mathbf{98.00}_{\pm2.00}$ & $\mathbf{96.00}_{\pm0.00}$ \\
\bottomrule
\end{tabularx}

\end{table*}

\par\addvspace{0.5\baselineskip}
\noindent\textbf{Motion disambiguation.} Benchmark (a) tests motion-direction ambiguity by pairing similar joint positions with opposing velocities, using replayed or newly generated action histories. Within 25 controls, \method{} achieves $96.67\%$ and $96.00\%$ success, compared with $75.33\%$ and $63.67\%$ for Visual history. Benchmark (b) isolates speed differences through 25 Slow--Fast pairs with near-identical positions, aligned velocity directions, and shared goals. On Slow cases, \method{} achieves $97.33\%$ within 25 controls, compared with $92.00\%$ for Visual history, rising to $100.00\%$ and $98.67\%$ within 50. Fast cases remain difficult within 25 controls, with $2.67\%$ success for \method{} and zero for the comparison models. Within 50 controls, \method{} reaches $96.00\%$, versus $57.33\%$ for Fast-LeWM and $58.67\%$ for Visual history. These results support executed-action context for planning under differences in motion direction and speed.

\par\addvspace{0.5\baselineskip}
\noindent\textbf{Goal-relative motion.} Benchmark (c) compares matched pairs of states moving toward or away from the goal. The away condition tests whether the controller can redirect ongoing motion toward the target. Within 25 controls, \method{} reaches $96.00\%$ in both conditions, while Fast-LeWM falls from $78.67\%$ toward the goal to $38.67\%$ away from it. \method{} reaches $100.00\%$ in both conditions within 50 controls. \method{} maintains high success whether the initial motion approaches or departs from the goal, suggesting that action history helps the controller account for ongoing motion when selecting goal-directed actions.

\par\addvspace{0.5\baselineskip}
\noindent\textbf{Goal adaptation.} Benchmark (d) changes the goal after 25 controls toward the original target. The executed history and current state remain fixed, while the goal displacement is retained or rotated by $90^\circ$ or $180^\circ$ in joint space at the same initial distance. Under a $180^\circ$ goal switch, \method{} maintains a 25-control success rate of $88.67\%$, close to its $89.33\%$ on the original goals. In contrast, Fast-LeWM's success rate falls from $88.00\%$ to $62.00\%$. Within 50 controls, \method{} reaches $98.67\%$ on reversed goals, compared with $94.00\%$ for Visual history. These results suggest that executed-action context supports control after a goal change, beyond simply continuing toward the original goal.

\par\addvspace{0.5\baselineskip}
\noindent\textbf{Overall results.} Across the four benchmarks, \method{} averages $83.08\%$ success within 25 controls and $99.11\%$ within 50, compared with $66.71\%$ and $90.32\%$ for Visual history. These gains suggest that executed-action memory captures motion context more effectively than the visual-history encoder. On average, the advantage is larger within 25 controls and remains substantial within 50 controls. These findings highlight executed actions as a useful source of dynamic context for motion disambiguation, goal-relative motion, and goal adaptation.

\subsection{Analysis of RAMP}
\label{sec:ramp_task_analysis}
\begin{figure}[t]
\centering
\includegraphics[width=\columnwidth]{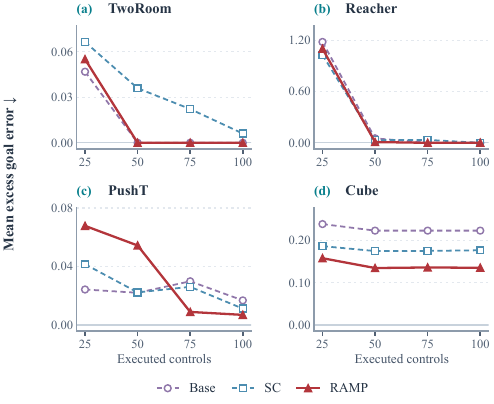}
\caption{\textbf{Goal error across execution budgets.} Mean excess error over three training seeds at 25, 50, 75, and 100 controls. Error is measured at first success or budget exhaustion.}
\label{fig:execution_budget}
\end{figure}

We compare Base, SC, and RAMP over 25--100 controls with all other settings fixed, as shown in Fig.~\ref{fig:execution_budget}. The metric is mean goal error exceeding the normalized success tolerance, evaluated when the task succeeds or the control budget is exhausted. This metric captures partial progress that binary success rates leave unreported. On Reacher, RAMP lowers the 50-control error to $0.009$, versus $0.052$ for Base and $0.031$ for SC, before all three reach zero by 100 controls. On Cube, RAMP maintains the lowest error across all budgets, reaching $0.135$ at 100 controls versus $0.223$ and $0.177$. On PushT, its advantage emerges at 75--100 controls, following higher errors at shorter budgets. RAMP also matches Base's zero error on TwoRoom from 50 controls onward. These results show that RAMP can improve goal-reaching accuracy under limited control budgets and reduce residual error with longer execution.

\begin{table}[t]
\centering\small
\setlength{\tabcolsep}{3pt}
\caption{\textbf{Planning time (seconds per CEM call).} Mean time per CEM call for batches of 50 environments under a 50-control execution budget, averaged across three training seeds. Bold and underline indicate the fastest and second-fastest results.}
\label{tab:planning_time}
\begin{tabularx}{\columnwidth}{@{}Xrrrr@{}}
\toprule
Model / Planner & TwoRoom & Reacher & PushT & Cube \\
\midrule
LeWM & 118.92 & 107.07 & 112.17 & 105.99 \\
Fast-LeWM & \underline{9.80} & \underline{10.05} & \underline{9.71} & \underline{11.19} \\
Fast-LeWM + SC & 14.97 & 16.66 & 14.74 & 16.28 \\
\midrule
\multicolumn{5}{l}{\textit{PACE-WM planning variants}} \\
\rowcolor{harprow}
Base & \textbf{9.44} & \textbf{9.30} & \textbf{9.03} & \textbf{10.64} \\
SC & 13.58 & 13.75 & 13.77 & 14.33 \\
\rowcolor{harprow}
RAMP & 13.43 & 14.38 & 13.61 & 14.20 \\
\bottomrule
\end{tabularx}

\end{table}

\subsection{Ablation Studies}
\label{sec:controlled_ablations}
\noindent\textbf{History representation.} We compare history encoders while fixing the parallel predictor architecture and RAMP planner, as reported in Table~\ref{tab:combined_ablation}(a). Action LSTM achieves the highest mean success across all four tasks. On Reacher, it reaches $98.00\%$, versus $84.00\%$ without history and $86.00\%$ with Visual history. Linear action conditioning improves TwoRoom and Reacher but reduces success on PushT and Cube relative to the no-history variant. Recurrent action encoders perform more consistently, with GRU approaching LSTM on PushT and Cube. The contrast suggests that accumulating action context over time is more effective than a simple linear correction. These results support recurrent encoding of executed actions as an effective source of dynamic context.

\suppressfloats[t]
\begin{figure}[t]
\centering
\includegraphics[width=\columnwidth]{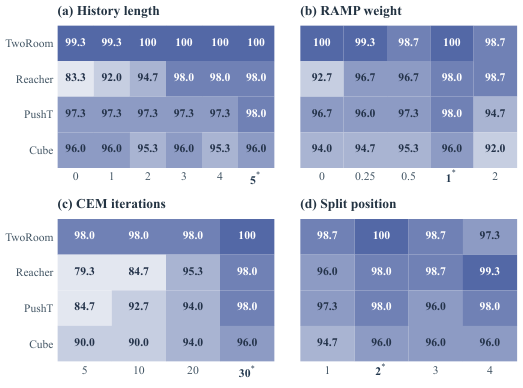}
\caption{\textbf{Parameter sensitivity.} Mean success rates (\%) over three training seeds when varying (a) retained history blocks $M$, (b) RAMP weight $\lambda$, (c) CEM iterations, and (d) split position $k$. Rows share the dataset labels on the left. Bold parameter labels with an asterisk mark the settings used in our main experiments.}
\label{fig:parameter_sensitivity}
\end{figure}

\noindent\textbf{Planning criterion.} We fix the action-LSTM model and vary only the planning criterion, as reported in Table~\ref{tab:combined_ablation}(b). Split-only uses the split prediction's terminal goal distance, while Unanchored minimizes $\max\{J_D,J_S\}$ without reference subtraction. RAMP exceeds SC and Split-only on all four tasks. It also improves over Unanchored on three tasks and matches its mean on PushT. On Cube, Unanchored reaches $92.00\%$, below Base at $94.00\%$, whereas RAMP achieves $96.00\%$. RAMP's advantage over Split-only and Unanchored supports combining both prediction paths through reference-relative progress.

\noindent\textbf{Complementary contributions.} Action memory improves Reacher success from $82.67\%$ to $92.67\%$ under Base planning, while RAMP further raises it to $98.00\%$. The complete model also matches or exceeds either component alone on the other tasks, supporting complementary benefits from executed-action context and the planning criterion.

\subsection{Planning Efficiency}
We measure planning time per CEM call for batches of 50 environments on an exclusively allocated GPU, using matched candidate and iteration budgets. As reported in Table~\ref{tab:planning_time}, \method{} with Base takes 9.03--10.64 seconds across the four tasks, providing approximately $10$--$13\times$ speedup over autoregressive LeWM and reducing planning time by $3.7$--$7.5\%$ relative to Fast-LeWM. RAMP adds computation for split prediction but retains approximately $7$--$9\times$ speedup over LeWM and remains faster than Fast-LeWM + SC on all four tasks. These results show that \method{} combines action-history conditioning with efficient parallel planning, while retaining a substantial speed advantage over autoregressive prediction even with RAMP.

\subsection{Parameter Sensitivity}
\label{sec:parameter_sensitivity}
To assess how key design and planning choices affect performance, we vary history length, RAMP weight, CEM iterations, and split position, as shown in Fig.~\ref{fig:parameter_sensitivity}.

\par\addvspace{0.5\baselineskip}
\noindent\textbf{History length.} With the trained model fixed, we vary the number of recent executed-action blocks used to reconstruct memory before planning. On Reacher, success rises from $83.33\%$ with reset memory to $98.00\%$ with three blocks. Three blocks match the full-history mean on three tasks and nearly match it on PushT, indicating that a short recent history recovers most of the observed benefit. The lower success rate with reset memory further supports the role of past actions in planning.

\par\addvspace{0.5\baselineskip}
\noindent\textbf{RAMP weight.} Increasing the penalty weight does not consistently improve performance, with $\lambda=2$ reducing success on PushT and Cube relative to $\lambda=1$. Unit weight achieves the best or joint-best mean on three tasks, supporting $\lambda=1$ as a shared setting that also admits the max--min interpretation in Eq.~\eqref{eq:maxmin}.

\par\addvspace{0.5\baselineskip}
\noindent\textbf{CEM iterations.} Increasing the search budget improves success most clearly on Reacher and PushT. Raising the iteration count from 5 to 30 increases their success rates from $79.33\%$ to $98.00\%$ and from $84.67\%$ to $98.00\%$, respectively. These gains highlight the value of iterative candidate refinement within parallel world-model planning.

\par\addvspace{0.5\baselineskip}
\noindent\textbf{Split position.} Success varies by at most $3.33$ percentage points within each task across the tested positions $k=1,\ldots,4$. The default $k=2$ achieves the best or joint-best mean on three tasks, while Reacher performs best at $k=4$. The modest variation suggests that RAMP does not depend on a narrowly tuned split position, with $k=2$ providing a competitive shared choice.

\section{Conclusion}
We present \method{}, a JEPA-style world model that combines executed-action memory with parallel latent prediction for visual planning.\ RAMP complements this architecture by evaluating reference-relative goal progress through direct and split predictions. Experiments across four tasks demonstrate improved planning success and substantially lower planning time than autoregressive LeWM. Controlled benchmarks and ablations show that action history helps interpret ongoing motion and adapt control to changing goals, supporting its role in efficient visual planning. Future work could explore more expressive action-memory models and adaptive planning criteria.

{\small\bibliographystyle{ieeenat_fullname}\bibliography{main_1}}

\clearpage
\raggedbottom
\section*{Appendix}
\setcounter{figure}{0}
\setcounter{table}{0}
\setcounter{equation}{0}
\setcounter{algorithm}{0}
\renewcommand{\thefigure}{S\arabic{figure}}
\renewcommand{\thetable}{S\arabic{table}}
\renewcommand{\theequation}{S\arabic{equation}}
\renewcommand{\thealgorithm}{S\arabic{algorithm}}
\renewcommand{\theHfigure}{appendix.\arabic{figure}}
\renewcommand{\theHtable}{appendix.\arabic{table}}
\renewcommand{\theHequation}{appendix.\arabic{equation}}
\providecommand{\theHalgorithm}{}
\renewcommand{\theHalgorithm}{appendix.\arabic{algorithm}}
\let\arxivOriginalLabel\label
\renewcommand{\label}[1]{\arxivOriginalLabel{#1}\arxivOriginalLabel{supp-#1}}
\section*{Overview}
This supplement provides implementation and evaluation details in Appendix~\ref{app:implementation}, mathematical analysis in Appendix~\ref{app:theory}, and benchmark construction with illustrative examples in Appendix~\ref{app:benchmarks}. Additional experiments and diagnostics appear in Appendix~\ref{app:experiments}.
\appendix
\section{Implementation and Evaluation Protocol}
\label{app:implementation}

\subsection{Model and Training}
\label{app:architecture}
\paragraph{Architecture.}
We use a ViT-Tiny visual encoder with $224\times224$ inputs, $14\times14$ patches, 12 layers, 3 attention heads, and width 192. Its CLS token is mapped to a 192-dimensional latent representation through a two-layer projection head with BatchNorm and GELU. The encoder is initialized without pretrained weights. The predictor uses sinusoidal positional encodings and a two-layer output head with LayerNorm and GELU, without a final residual addition of the current embedding.

\par\addvspace{0.5\baselineskip}
\noindent\textbf{Optimization.}
Training supervises the predicted latent representations after each action block, with prediction horizons ranging from 1 to 5 blocks. We follow the task-specific batch-size settings of Fast-LeWM~\cite{gao2026fastlewm} and use three independent training seeds per configuration.

\subsection{Planning Settings}
\label{app:planning}
We summarize the RAMP planning procedure, including reference insertion, candidate-specific memory updates, and closed-loop execution, in Algorithm~\ref{alg:ramp_supp}. Planner comparisons use identical checkpoints and evaluation cases within each task and training seed. Timing experiments run exclusively on a single GPU.

\begin{algorithm}[t]
\caption{CEM Planning with RAMP}
\label{alg:ramp_supp}
\begin{algorithmic}[1]
\Require Image $o_t$, goal $o_g$, action memory $m_t$
\State Encode $z_t,z_g$ and initialize the CEM distribution
\For{each CEM iteration}
 \State Sample bounded candidates and insert $A_{\mathrm{ref}}=\mu$
 \ForAll{candidates $A$ in parallel}
  \State Predict $\widehat Z=F_\theta(z_t,A;m_t)$
  \State Set $z_D=\widehat z_H$ and split anchor $\widehat z_k$
  \State Advance a copy of $m_t$ using original $A_{1:k}$
  \State Predict the suffix from $\widehat z_k$ to obtain $z_S$
  \State Compute direct and split goal costs $J_D,J_S$
 \EndFor
 \ForAll{candidates $A$ in parallel}
  \State $\Delta_D(A)\gets J_D(A_{\mathrm{ref}})-J_D(A)$
  \State $\Delta_S(A)\gets J_S(A_{\mathrm{ref}})-J_S(A)$
  \State $P(A)\gets[\Delta_D(A)-\Delta_S(A)]_+$
 \EndFor
 \State Rank candidates by $J_D(A)+\lambda P(A)$
 \State Update the sampling distribution from elites
\EndFor
\State Execute the final mean and update real action memory
\State Replan from the new observation if budget remains
\end{algorithmic}
\end{algorithm}

\section{Mathematical Analysis}
\label{app:theory}

\label{app:ramp_properties}
\paragraph{Proof of the max--min interpretation.}
Fix a reference action sequence $A_{\rm ref}$ within a CEM iteration and set $\lambda=1$. Substituting $J_D(A)=J_D(A_{\rm ref})-\Delta_D(A)$ into the RAMP objective gives:
\begin{equation}
\begin{aligned}
 J_{\rgp}(A)
 &=J_D(A_{\rm ref})-\Delta_D(A)\\
 &\quad+[\Delta_D(A)-\Delta_S(A)]_+.
\end{aligned}
\end{equation}
We distinguish two cases based on predicted progress:

\noindent\textbf{Case 1:} $\Delta_D(A)\geq\Delta_S(A)$. The penalty is $\Delta_D(A)-\Delta_S(A)$, so the objective becomes:
\begin{equation}
 J_{\rgp}(A)=J_D(A_{\rm ref})-\Delta_S(A).
\end{equation}
\noindent\textbf{Case 2:} $\Delta_D(A)<\Delta_S(A)$. The penalty is zero, so the objective becomes:
\begin{equation}
 J_{\rgp}(A)=J_D(A_{\rm ref})-\Delta_D(A).
\end{equation}
Combining the two cases yields:
\begin{equation}
 J_{\rgp}(A)=J_D(A_{\rm ref})-\min\{\Delta_D(A),\Delta_S(A)\}.
\end{equation}
The reference cost $J_D(A_{\rm ref})$ is constant across candidates within the iteration. Minimizing the RAMP objective therefore maximizes the smaller of the two predicted progress values. This equivalence concerns model predictions and does not guarantee progress in the environment.

\par\addvspace{0.5\baselineskip}
\noindent\textbf{Invariance to additive offsets.}
For candidate-independent offsets $J'_p(A)=J_p(A)+b_p$, $p\in\{D,S\}$, reference subtraction gives $\Delta'_p=\Delta_p$. Thus $J'_{\rgp}(A)=J_{\rgp}(A)+b_D$, preserving candidate ordering for any fixed $\lambda$. A common positive scaling also preserves ordering, whereas separate path scalings or candidate-dependent offsets need not.

\par\addvspace{0.5\baselineskip}
\noindent\textbf{Conditional candidate selection.}
At $\lambda=1$, if the reference belongs to the scored candidate set, its progress is zero under both paths. A minimum-cost scored candidate therefore has $\min(\Delta_D,\Delta_S)\geq0$. This property applies to scored-candidate selection, not automatically to the final updated mean returned by CEM.

\section{Benchmark Construction}
\label{app:benchmarks}

\label{app:benchmark_protocol}
\label{app:qualitative}
\label{app:motion_examples}
\label{sec:motion_examples}
\par\addvspace{0.5\baselineskip}
\noindent\textbf{Shared evaluation protocol.}
We keep each model's training configuration fixed across the four benchmarks and use Base planning for evaluation. Under the unified protocol, history-conditioned models receive the 25 most recent actions or the corresponding visual history in all four benchmarks. Model evaluation begins from the state reached after executing the historical actions. We report success within 25 and 50 steps from the same closed-loop run. Evaluation cases are fixed for each benchmark before its corresponding model evaluations. We describe benchmark construction below using Reacher as an example. To assess the applicability of our benchmark design and model across tasks, we extend these benchmarks to Cube and report the results in Appendix~\ref{app:cube_benchmarks}.

\begin{figure}[!tp]
\centering
\includegraphics[width=.92\columnwidth]{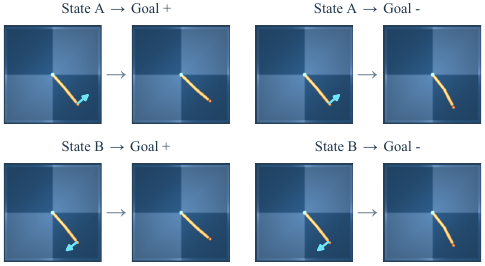}
\caption{\textbf{Four state--goal combinations in the direction-ambiguity benchmark.} States A and B have similar joint configurations but opposing motion directions. Each is evaluated against Goal $+$ and Goal $-$, yielding four scenes per pair and 100 scenes for 25 pairs.}
\label{fig:motion_speed_teaser}
\par\smallskip
\centering
\includegraphics[width=.92\columnwidth]{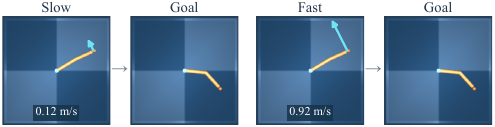}
\caption{\textbf{Speed ambiguity.} Similar configurations with aligned motion directions but different speeds.}
\label{fig:benchmark_b_example}
\par\smallskip
\centering
\includegraphics[width=.92\columnwidth]{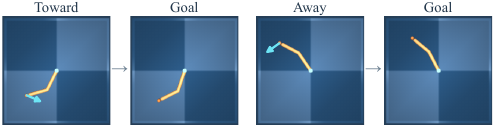}
\caption{\textbf{Goal-relative motion.} Matched Toward and Away cases with their respective goals shown side by side.}
\label{fig:benchmark_d_example}
\par\smallskip
\centering
\includegraphics[width=.92\columnwidth]{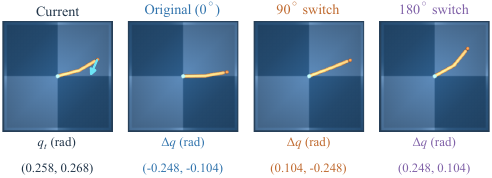}
\caption{\textbf{Goal switching.} The current state and action history remain fixed while the goal displacement $\Delta q=q_g-q_t$ is rotated in joint space. Values below the images show the current joint configuration and the three goal displacements in radians.}
\label{fig:goal_switch_example}
\end{figure}

\par\addvspace{0.5\baselineskip}
\noindent\textbf{(a) Direction ambiguity.}
We form 25 state pairs with near-identical joint positions and opposing velocities. Two shared goals per pair yield 100 scenes, as illustrated in Fig.~\ref{fig:motion_speed_teaser}. Histories are either replayed from recorded trajectories or generated by executing actions sampled from $\mathcal{U}([-1,1]^2)$, using the same pairing criteria. Generated histories start from source states and yield new handover states to test action-history generalization.

\par\addvspace{0.5\baselineskip}
\noindent\textbf{(b) Speed ambiguity.} %
\label{app:speed}%
\label{sec:speed_magnitude}%
We form 25 pairs of slow and fast states with similar joint configurations, aligned motion directions, and a shared goal, as shown in Fig.~\ref{fig:benchmark_b_example}. We vary initial positions and pulse actions in the native simulator to generate slow and fast motion. Slow fingertip speeds range from $0.112$ to $0.127$ m/s and Fast speeds from $0.919$ to $0.991$ m/s. The Slow condition is evaluated separately on all 25 states, while Fast results are retained from the unchanged cases.

\noindent\textbf{(c) Goal-relative motion.}
We construct 25 Toward--Away pairs, totaling 50 scenes. States are classified by the sign of $(q_t-q_g)^\top\dot q_t$, where $q_t$, $q_g$, and $\dot q_t$ denote the current joint configuration, target configuration, and joint velocity. Negative values indicate motion toward the goal, while nonnegative values indicate non-approaching motion. We match pairs by joint-speed magnitude, goal distance, and reference policy, as illustrated in Fig.~\ref{fig:benchmark_d_example}.

\par\addvspace{0.5\baselineskip}
\noindent\textbf{(d) Goal switching.}
We construct 50 histories by executing 25 actions with a reference policy. At the resulting state, we retain the original goal or rotate its joint-space displacement by $90^\circ$ or $180^\circ$, yielding three goal conditions per history. Specifically, $q_g^{(\theta)}=q_t+R_\theta(q_g-q_t)$, where $R_\theta$ is the rotation matrix. This preserves the history, current state, and initial goal distance while changing the required motion direction, as illustrated in Fig.~\ref{fig:goal_switch_example}.

\section{Additional Experiments and Diagnostics}
\label{app:experiments}

\subsection{Reproduced Baselines under Matched Settings}
\label{app:matched_baselines}
\begin{table}[!t]
\centering\footnotesize
\setlength{\tabcolsep}{2pt}
\caption{\textbf{Reproduced baselines under matched planning settings.} Success rates (\%) are reported as means across three training seeds. Avg. is the unweighted four-task mean. Bold and underline indicate the best and second-best means.}
\label{tab:matched_baseline_reproduction}
\begin{tabularx}{\columnwidth}{@{}Xrrrrr@{}}
\toprule
Model & TwoRoom & Reacher & PushT & Cube & Avg. \\
\midrule
LeWM~\cite{maes2026lewm} & $89.33$ & $86.00$ & $92.00$ & $82.67$ & $87.50$\\
Fast-LeWM~\cite{gao2026fastlewm} & $\underline{93.33}$ & $78.67$ & $90.67$ & $91.33$ & $88.50$\\
Fast-LeWM + SC~\cite{gao2026fastlewm} & $92.00$ & $78.67$ & $91.33$ & $92.00$ & $88.50$\\
\midrule
\rowcolor{harprow}
\method{} & $\mathbf{100.00}$ & $\underline{92.67}$ & $\underline{96.67}$ & $\underline{94.00}$ & $\underline{95.83}$\\
\rowcolor{harprow}
\method{} + RAMP & $\mathbf{100.00}$ & $\mathbf{98.00}$ & $\mathbf{98.00}$ & $\mathbf{96.00}$ & $\mathbf{98.00}$\\
\bottomrule
\end{tabularx}

\end{table}

To support a fair comparison, we reproduce LeWM, Fast-LeWM, and Fast-LeWM + SC under training and evaluation settings matched to PACE-WM. Our full method achieves a four-task average success rate of $98.00\%$, compared with $87.50\%$ for LeWM and $88.50\%$ for both Fast-LeWM variants, as reported in Table~\ref{tab:matched_baseline_reproduction}. It outperforms all reproduced baselines on each task, demonstrating stronger visual planning performance under matched settings.

\begin{figure}[H]
\centering
\includegraphics[width=\columnwidth]{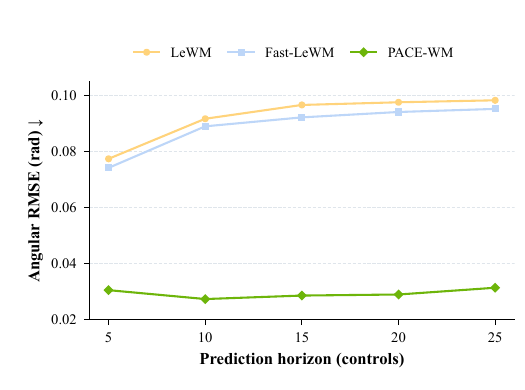}
\caption{\textbf{Prediction error across horizons.} Curves show mean angular RMSE (rad) obtained through model-specific linear probes over three training seeds.}
\label{fig:prediction_horizon}
\end{figure}

\subsection{Prediction Error across Horizons}
\label{app:prediction}
\label{sec:prediction_horizon}
We compare LeWM, Fast-LeWM, and PACE-WM on the same 256 Reacher trajectories, evaluating horizons of 5--25 steps across three training seeds. For each model, a linear probe predicts the sine and cosine of joint angles from latent representations, and angles are recovered using $\operatorname{atan2}$. Probes are fitted only on real-image embeddings, using separate trajectories for probe fitting and evaluation. We report angular RMSE in radians.

Fast-LeWM exhibits lower prediction error than LeWM across the evaluated horizons, as shown in Fig.~\ref{fig:prediction_horizon}. PACE-WM further reduces the error and maintains a relatively stable profile as the horizon increases. At 25 steps, PACE-WM achieves an RMSE of $0.031$ rad, compared with $0.098$ for LeWM and $0.095$ for Fast-LeWM. These results support PACE-WM's history-conditioned parallel architecture for predicting future joint configurations.

\subsection{Action-Memory Analysis}
\label{app:history}
\label{app:association}
\label{sec:mechanism}
\begin{table}[t]
\centering\small
\setlength{\tabcolsep}{3pt}
\caption{\textbf{Action-memory interventions.} Future-association accuracy (\%) on 25 state pairs using 25 past actions, reported as mean $\pm$ SD across three training seeds. Best and second-best means are bold and underlined, respectively.}
\label{tab:history_v2}
\begin{tabular*}{\columnwidth}{@{\extracolsep{\fill}}lrrr@{}}\toprule
Action history & 5 steps & 15 steps & 25 steps\\\midrule
Correct & $\mathbf{94.00}{\scriptstyle\,\pm 2.00}$ & $\mathbf{94.67}{\scriptstyle\,\pm 1.15}$ & $\mathbf{92.67}{\scriptstyle\,\pm 3.06}$\\
Removed & $\underline{50.67}{\scriptstyle\,\pm 1.15}$ & $\underline{52.00}{\scriptstyle\,\pm 0.00}$ & $\underline{50.00}{\scriptstyle\,\pm 0.00}$\\
Swapped & $2.67{\scriptstyle\,\pm 1.15}$ & $2.00{\scriptstyle\,\pm 0.00}$ & $2.00{\scriptstyle\,\pm 0.00}$\\
Reversed & $\underline{50.67}{\scriptstyle\,\pm 1.15}$ & $48.67{\scriptstyle\,\pm 1.15}$ & $47.33{\scriptstyle\,\pm 1.15}$\\
\bottomrule\end{tabular*}

\end{table}

\begin{table}[t]
\centering\small
\setlength{\tabcolsep}{3pt}
\caption{\textbf{Robustness to imperfect action history.} Future-association accuracy (\%) on the same 25 state pairs. Perturbed results average three fixed corruption draws per model before reporting mean $\pm$ SD across three training seeds. Best and second-best means are bold and underlined, respectively.}
\label{tab:history_robustness}
\begin{tabular*}{\columnwidth}{@{\extracolsep{\fill}}lrrr@{}}\toprule
Action history & 5 steps & 15 steps & 25 steps\\\midrule
Correct & $\mathbf{94.00}{\scriptstyle\,\pm 2.00}$ & $\mathbf{94.67}{\scriptstyle\,\pm 1.15}$ & $\mathbf{92.67}{\scriptstyle\,\pm 3.06}$\\
Noisy & $\underline{93.78}{\scriptstyle\,\pm 1.02}$ & $\underline{94.00}{\scriptstyle\,\pm 0.00}$ & $\mathbf{92.67}{\scriptstyle\,\pm 1.76}$\\
Missing & $90.89{\scriptstyle\,\pm 0.77}$ & $88.22{\scriptstyle\,\pm 1.92}$ & $\underline{87.56}{\scriptstyle\,\pm 1.68}$\\
\bottomrule\end{tabular*}

\end{table}

\begin{table*}[t]
\centering
\small
\setlength{\tabcolsep}{3pt}
\renewcommand{\arraystretch}{1.02}
\setlength{\aboverulesep}{1.5pt}
\setlength{\belowrulesep}{1.5pt}
\setlength{\arrayrulewidth}{0.3pt}
\caption{\textbf{Visual planning on four controlled Cube benchmarks.} Success rates (\%, mean $\pm$ sample SD over three training seeds). Avg. weights benchmarks equally after averaging their conditions. Bold/underline mark the best/second-best results per row.}
\label{tab:cube_history_benchmarks}
\begin{tabularx}{\textwidth}{@{}l|lc|*{5}{>{\raggedleft\arraybackslash}X}@{}}
\toprule
\textbf{Benchmark} & \textbf{Condition} & \textbf{Controls} & \multicolumn{1}{c}{LeWM} & \multicolumn{1}{c}{Fast-LeWM} & \multicolumn{1}{c}{w/o history} & \multicolumn{1}{c}{Visual history} & \multicolumn{1}{c}{\cellcolor{harprow}\method{}} \\

\midrule
\multirow{2}{*}{\textbf{(a) Direction ambiguity}} & Generated history & 25 & \makebox[2.9em][r]{13.33}{\scriptsize\makebox[0.9em][r]{$\pm$}\makebox[2.6em][l]{8.50}} & \makebox[2.9em][r]{50.33}{\scriptsize\makebox[0.9em][r]{$\pm$}\makebox[2.6em][l]{12.34}} & \makebox[2.9em][r]{51.00}{\scriptsize\makebox[0.9em][r]{$\pm$}\makebox[2.6em][l]{1.73}} & \makebox[2.9em][r]{\underline{52.67}}{\scriptsize\makebox[0.9em][r]{$\pm$}\makebox[2.6em][l]{4.62}} & \cellcolor{harprow}\makebox[2.9em][r]{\textbf{64.67}}{\scriptsize\makebox[0.9em][r]{$\pm$}\makebox[2.6em][l]{7.23}} \\
 & Generated history & 50 & \makebox[2.9em][r]{13.33}{\scriptsize\makebox[0.9em][r]{$\pm$}\makebox[2.6em][l]{8.50}} & \makebox[2.9em][r]{50.33}{\scriptsize\makebox[0.9em][r]{$\pm$}\makebox[2.6em][l]{12.34}} & \makebox[2.9em][r]{51.67}{\scriptsize\makebox[0.9em][r]{$\pm$}\makebox[2.6em][l]{2.89}} & \makebox[2.9em][r]{\underline{52.67}}{\scriptsize\makebox[0.9em][r]{$\pm$}\makebox[2.6em][l]{4.62}} & \cellcolor{harprow}\makebox[2.9em][r]{\textbf{65.67}}{\scriptsize\makebox[0.9em][r]{$\pm$}\makebox[2.6em][l]{6.66}} \\
\midrule
\multirow{4}{*}{\textbf{(b) Speed ambiguity}} & Slow & 25 & \makebox[2.9em][r]{82.67}{\scriptsize\makebox[0.9em][r]{$\pm$}\makebox[2.6em][l]{2.31}} & \makebox[2.9em][r]{\underline{90.67}}{\scriptsize\makebox[0.9em][r]{$\pm$}\makebox[2.6em][l]{2.31}} & \makebox[2.9em][r]{86.67}{\scriptsize\makebox[0.9em][r]{$\pm$}\makebox[2.6em][l]{2.31}} & \makebox[2.9em][r]{89.33}{\scriptsize\makebox[0.9em][r]{$\pm$}\makebox[2.6em][l]{2.31}} & \cellcolor{harprow}\makebox[2.9em][r]{\textbf{93.33}}{\scriptsize\makebox[0.9em][r]{$\pm$}\makebox[2.6em][l]{2.31}} \\
 & Fast & 25 & \makebox[2.9em][r]{84.00}{\scriptsize\makebox[0.9em][r]{$\pm$}\makebox[2.6em][l]{4.00}} & \makebox[2.9em][r]{85.33}{\scriptsize\makebox[0.9em][r]{$\pm$}\makebox[2.6em][l]{6.11}} & \makebox[2.9em][r]{\underline{86.67}}{\scriptsize\makebox[0.9em][r]{$\pm$}\makebox[2.6em][l]{2.31}} & \makebox[2.9em][r]{84.00}{\scriptsize\makebox[0.9em][r]{$\pm$}\makebox[2.6em][l]{4.00}} & \cellcolor{harprow}\makebox[2.9em][r]{\textbf{88.00}}{\scriptsize\makebox[0.9em][r]{$\pm$}\makebox[2.6em][l]{4.00}} \\
\cmidrule(l){2-8}
 & Slow & 50 & \makebox[2.9em][r]{84.00}{\scriptsize\makebox[0.9em][r]{$\pm$}\makebox[2.6em][l]{4.00}} & \makebox[2.9em][r]{\underline{92.00}}{\scriptsize\makebox[0.9em][r]{$\pm$}\makebox[2.6em][l]{0.00}} & \makebox[2.9em][r]{88.00}{\scriptsize\makebox[0.9em][r]{$\pm$}\makebox[2.6em][l]{0.00}} & \makebox[2.9em][r]{89.33}{\scriptsize\makebox[0.9em][r]{$\pm$}\makebox[2.6em][l]{2.31}} & \cellcolor{harprow}\makebox[2.9em][r]{\textbf{93.33}}{\scriptsize\makebox[0.9em][r]{$\pm$}\makebox[2.6em][l]{2.31}} \\
 & Fast & 50 & \makebox[2.9em][r]{84.00}{\scriptsize\makebox[0.9em][r]{$\pm$}\makebox[2.6em][l]{4.00}} & \makebox[2.9em][r]{86.67}{\scriptsize\makebox[0.9em][r]{$\pm$}\makebox[2.6em][l]{4.62}} & \makebox[2.9em][r]{\underline{88.00}}{\scriptsize\makebox[0.9em][r]{$\pm$}\makebox[2.6em][l]{0.00}} & \makebox[2.9em][r]{86.67}{\scriptsize\makebox[0.9em][r]{$\pm$}\makebox[2.6em][l]{2.31}} & \cellcolor{harprow}\makebox[2.9em][r]{\textbf{89.33}}{\scriptsize\makebox[0.9em][r]{$\pm$}\makebox[2.6em][l]{2.31}} \\
\midrule
\multirow{4}{*}{\textbf{(c) Goal-relative motion}} & Toward & 25 & \makebox[2.9em][r]{0.00}{\scriptsize\makebox[0.9em][r]{$\pm$}\makebox[2.6em][l]{0.00}} & \makebox[2.9em][r]{2.67}{\scriptsize\makebox[0.9em][r]{$\pm$}\makebox[2.6em][l]{2.31}} & \makebox[2.9em][r]{61.33}{\scriptsize\makebox[0.9em][r]{$\pm$}\makebox[2.6em][l]{8.33}} & \makebox[2.9em][r]{\underline{72.00}}{\scriptsize\makebox[0.9em][r]{$\pm$}\makebox[2.6em][l]{26.23}} & \cellcolor{harprow}\makebox[2.9em][r]{\textbf{86.67}}{\scriptsize\makebox[0.9em][r]{$\pm$}\makebox[2.6em][l]{16.17}} \\
 & Away & 25 & \makebox[2.9em][r]{0.00}{\scriptsize\makebox[0.9em][r]{$\pm$}\makebox[2.6em][l]{0.00}} & \makebox[2.9em][r]{0.00}{\scriptsize\makebox[0.9em][r]{$\pm$}\makebox[2.6em][l]{0.00}} & \makebox[2.9em][r]{73.33}{\scriptsize\makebox[0.9em][r]{$\pm$}\makebox[2.6em][l]{16.17}} & \makebox[2.9em][r]{\underline{77.33}}{\scriptsize\makebox[0.9em][r]{$\pm$}\makebox[2.6em][l]{25.72}} & \cellcolor{harprow}\makebox[2.9em][r]{\textbf{86.67}}{\scriptsize\makebox[0.9em][r]{$\pm$}\makebox[2.6em][l]{16.17}} \\
\cmidrule(l){2-8}
 & Toward & 50 & \makebox[2.9em][r]{0.00}{\scriptsize\makebox[0.9em][r]{$\pm$}\makebox[2.6em][l]{0.00}} & \makebox[2.9em][r]{2.67}{\scriptsize\makebox[0.9em][r]{$\pm$}\makebox[2.6em][l]{2.31}} & \makebox[2.9em][r]{61.33}{\scriptsize\makebox[0.9em][r]{$\pm$}\makebox[2.6em][l]{8.33}} & \makebox[2.9em][r]{\underline{72.00}}{\scriptsize\makebox[0.9em][r]{$\pm$}\makebox[2.6em][l]{26.23}} & \cellcolor{harprow}\makebox[2.9em][r]{\textbf{86.67}}{\scriptsize\makebox[0.9em][r]{$\pm$}\makebox[2.6em][l]{16.17}} \\
 & Away & 50 & \makebox[2.9em][r]{4.00}{\scriptsize\makebox[0.9em][r]{$\pm$}\makebox[2.6em][l]{4.00}} & \makebox[2.9em][r]{0.00}{\scriptsize\makebox[0.9em][r]{$\pm$}\makebox[2.6em][l]{0.00}} & \makebox[2.9em][r]{73.33}{\scriptsize\makebox[0.9em][r]{$\pm$}\makebox[2.6em][l]{16.17}} & \makebox[2.9em][r]{\underline{77.33}}{\scriptsize\makebox[0.9em][r]{$\pm$}\makebox[2.6em][l]{25.72}} & \cellcolor{harprow}\makebox[2.9em][r]{\textbf{90.67}}{\scriptsize\makebox[0.9em][r]{$\pm$}\makebox[2.6em][l]{9.24}} \\
\midrule
\multirow{6}{*}{\textbf{(d) Goal switching}} & $0^\circ$ (original) & 25 & \makebox[2.9em][r]{55.33}{\scriptsize\makebox[0.9em][r]{$\pm$}\makebox[2.6em][l]{13.32}} & \makebox[2.9em][r]{41.33}{\scriptsize\makebox[0.9em][r]{$\pm$}\makebox[2.6em][l]{6.43}} & \makebox[2.9em][r]{\underline{62.67}}{\scriptsize\makebox[0.9em][r]{$\pm$}\makebox[2.6em][l]{15.01}} & \makebox[2.9em][r]{60.67}{\scriptsize\makebox[0.9em][r]{$\pm$}\makebox[2.6em][l]{8.08}} & \cellcolor{harprow}\makebox[2.9em][r]{\textbf{76.67}}{\scriptsize\makebox[0.9em][r]{$\pm$}\makebox[2.6em][l]{4.16}} \\
 & $90^\circ$ switch & 25 & \makebox[2.9em][r]{34.00}{\scriptsize\makebox[0.9em][r]{$\pm$}\makebox[2.6em][l]{5.29}} & \makebox[2.9em][r]{37.33}{\scriptsize\makebox[0.9em][r]{$\pm$}\makebox[2.6em][l]{9.02}} & \makebox[2.9em][r]{50.67}{\scriptsize\makebox[0.9em][r]{$\pm$}\makebox[2.6em][l]{7.02}} & \makebox[2.9em][r]{\underline{53.33}}{\scriptsize\makebox[0.9em][r]{$\pm$}\makebox[2.6em][l]{8.08}} & \cellcolor{harprow}\makebox[2.9em][r]{\textbf{60.00}}{\scriptsize\makebox[0.9em][r]{$\pm$}\makebox[2.6em][l]{12.00}} \\
 & $180^\circ$ switch & 25 & \makebox[2.9em][r]{32.67}{\scriptsize\makebox[0.9em][r]{$\pm$}\makebox[2.6em][l]{3.06}} & \makebox[2.9em][r]{23.33}{\scriptsize\makebox[0.9em][r]{$\pm$}\makebox[2.6em][l]{11.55}} & \makebox[2.9em][r]{\underline{52.00}}{\scriptsize\makebox[0.9em][r]{$\pm$}\makebox[2.6em][l]{10.00}} & \makebox[2.9em][r]{49.33}{\scriptsize\makebox[0.9em][r]{$\pm$}\makebox[2.6em][l]{9.24}} & \cellcolor{harprow}\makebox[2.9em][r]{\textbf{54.67}}{\scriptsize\makebox[0.9em][r]{$\pm$}\makebox[2.6em][l]{3.06}} \\
\cmidrule(l){2-8}
 & $0^\circ$ (original) & 50 & \makebox[2.9em][r]{56.00}{\scriptsize\makebox[0.9em][r]{$\pm$}\makebox[2.6em][l]{13.86}} & \makebox[2.9em][r]{41.33}{\scriptsize\makebox[0.9em][r]{$\pm$}\makebox[2.6em][l]{6.43}} & \makebox[2.9em][r]{\underline{62.67}}{\scriptsize\makebox[0.9em][r]{$\pm$}\makebox[2.6em][l]{15.01}} & \makebox[2.9em][r]{60.67}{\scriptsize\makebox[0.9em][r]{$\pm$}\makebox[2.6em][l]{8.08}} & \cellcolor{harprow}\makebox[2.9em][r]{\textbf{77.33}}{\scriptsize\makebox[0.9em][r]{$\pm$}\makebox[2.6em][l]{3.06}} \\
 & $90^\circ$ switch & 50 & \makebox[2.9em][r]{35.33}{\scriptsize\makebox[0.9em][r]{$\pm$}\makebox[2.6em][l]{6.43}} & \makebox[2.9em][r]{37.33}{\scriptsize\makebox[0.9em][r]{$\pm$}\makebox[2.6em][l]{9.02}} & \makebox[2.9em][r]{50.67}{\scriptsize\makebox[0.9em][r]{$\pm$}\makebox[2.6em][l]{7.02}} & \makebox[2.9em][r]{\underline{53.33}}{\scriptsize\makebox[0.9em][r]{$\pm$}\makebox[2.6em][l]{8.08}} & \cellcolor{harprow}\makebox[2.9em][r]{\textbf{60.67}}{\scriptsize\makebox[0.9em][r]{$\pm$}\makebox[2.6em][l]{11.02}} \\
 & $180^\circ$ switch & 50 & \makebox[2.9em][r]{33.33}{\scriptsize\makebox[0.9em][r]{$\pm$}\makebox[2.6em][l]{2.31}} & \makebox[2.9em][r]{23.33}{\scriptsize\makebox[0.9em][r]{$\pm$}\makebox[2.6em][l]{11.55}} & \makebox[2.9em][r]{\underline{52.67}}{\scriptsize\makebox[0.9em][r]{$\pm$}\makebox[2.6em][l]{9.02}} & \makebox[2.9em][r]{49.33}{\scriptsize\makebox[0.9em][r]{$\pm$}\makebox[2.6em][l]{9.24}} & \cellcolor{harprow}\makebox[2.9em][r]{\textbf{54.67}}{\scriptsize\makebox[0.9em][r]{$\pm$}\makebox[2.6em][l]{3.06}} \\
\midrule
\multicolumn{2}{c}{} & 25 & \makebox[2.9em][r]{34.33}{\scriptsize\phantom{\makebox[0.9em][r]{$\pm$}\makebox[2.6em][l]{0.00}}} & \makebox[2.9em][r]{43.42}{\scriptsize\phantom{\makebox[0.9em][r]{$\pm$}\makebox[2.6em][l]{0.00}}} & \makebox[2.9em][r]{65.03}{\scriptsize\phantom{\makebox[0.9em][r]{$\pm$}\makebox[2.6em][l]{0.00}}} & \makebox[2.9em][r]{\underline{67.11}}{\scriptsize\phantom{\makebox[0.9em][r]{$\pm$}\makebox[2.6em][l]{0.00}}} & \cellcolor{harprow}\makebox[2.9em][r]{\textbf{76.44}}{\scriptsize\phantom{\makebox[0.9em][r]{$\pm$}\makebox[2.6em][l]{0.00}}} \\
\multicolumn{2}{c}{\multirow{-2}{*}{\textbf{Average over four benchmarks}}} & 50 & \makebox[2.9em][r]{35.22}{\scriptsize\phantom{\makebox[0.9em][r]{$\pm$}\makebox[2.6em][l]{0.00}}} & \makebox[2.9em][r]{43.75}{\scriptsize\phantom{\makebox[0.9em][r]{$\pm$}\makebox[2.6em][l]{0.00}}} & \makebox[2.9em][r]{65.58}{\scriptsize\phantom{\makebox[0.9em][r]{$\pm$}\makebox[2.6em][l]{0.00}}} & \makebox[2.9em][r]{\underline{67.44}}{\scriptsize\phantom{\makebox[0.9em][r]{$\pm$}\makebox[2.6em][l]{0.00}}} & \cellcolor{harprow}\makebox[2.9em][r]{\textbf{77.47}}{\scriptsize\phantom{\makebox[0.9em][r]{$\pm$}\makebox[2.6em][l]{0.00}}} \\
\bottomrule
\end{tabularx}

\end{table*}

\noindent\textbf{History interventions.} We test PACE-WM's dependence on action history using 25 pairs of visually similar states with different velocities sampled from 1,024 Reacher trajectories. Source trajectories may overlap training data. Each state has 25 past actions, and paired states receive identical future actions from a third trajectory. With model weights and other inputs fixed, we keep the history, remove its correction, swap histories within each pair, or reverse their order. Future-association accuracy measures whether a predicted embedding is closer to its own observed future than to the paired alternative. At 25 steps, correct history achieves $92.67\%$, compared with $50.00\%$ after removal, $2.00\%$ after swapping, and $47.33\%$ after reversal, as reported in Table~\ref{tab:history_v2}. These changes indicate that predictions depend on both the content and temporal order of action history.

\noindent\textbf{History robustness.} We assess whether PACE-WM retains its predictive advantage when recorded action histories are noisy or incomplete. Using the same states and trained models, we perturb only the recorded past actions. We add independent Gaussian noise with a standard deviation of $5\%$ of each action dimension's training standard deviation, or randomly mask two of the 25 past actions. Missing positions retain their temporal slots and use zero-valued inputs with validity flags set to zero. Each model is evaluated with the same three random perturbations, and results are averaged before aggregation across training seeds. At 25 steps, noisy history retains $92.67\%$ accuracy, while missing actions reduce it to $87.56\%$, as reported in Table~\ref{tab:history_robustness}. Both remain above the $50.00\%$ obtained after history removal. These results suggest that PACE-WM tolerates mild noise in recorded actions and retains useful historical context despite partial action loss.

\subsection{Controlled Cube Benchmarks}
\label{app:cube_benchmarks}
We extend the four benchmarks to Cube using the same model variants and evaluation protocol. PACE-WM achieves the highest mean success rate in all conditions at both control budgets in Table~\ref{tab:cube_history_benchmarks}. Its four-benchmark averages reach $76.44\%$ within 25 controls and $77.47\%$ within 50. The gains over Visual history are 9.33 and 10.03 percentage points, respectively.

The gains vary across motion conditions. On direction ambiguity, PACE-WM exceeds Visual history by 12.00 points within 25 controls and 13.00 within 50. This supports the use of executed actions to distinguish motion states with similar appearances. On speed ambiguity, PACE-WM achieves $93.33\%$ success on Slow cases and $89.33\%$ on Fast cases within 50 controls. The strongest alternatives achieve $92.00\%$ and $88.00\%$. On goal-relative motion, its 25-control Toward success reaches $86.67\%$. The no-history and visual-history variants achieve $61.33\%$ and $72.00\%$. On Away cases, PACE-WM improves from $86.67\%$ to $90.67\%$ with additional controls. This suggests that action context can aid recovery when ongoing motion does not approach the goal. Goal switching remains more challenging. PACE-WM reaches $77.33\%$ on original goals within 50 controls. Success falls to $60.67\%$ for $90^\circ$ switches and $54.67\%$ for $180^\circ$ switches. PACE-WM maintains the highest success rates under both goal switches, supporting its ability to adapt action selection to changed goals.

Across the four Cube benchmarks, PACE-WM achieves the highest mean success rates at both control budgets. These results extend the evidence for executed-action memory beyond Reacher and support its value as dynamic context for visual planning across tasks.

\end{document}